\documentclass[letterpaper]{article} 
\usepackage{aaai25}  
\usepackage{times}  
\usepackage{helvet}  
\usepackage{courier}  
\usepackage[hyphens]{url}  
\usepackage{graphicx} 
\usepackage{tablefootnote}
\usepackage{natbib}  
\usepackage{caption} 
\usepackage{lipsum}
\usepackage[disable]{todonotes}

\usepackage{amsthm}
\usepackage{amsmath}
\usepackage{amsfonts}
\usepackage{amssymb}

\theoremstyle{plain}

\theoremstyle{definition}

\theoremstyle{remark}

\usepackage{algpseudocodex}
\usepackage{algorithm}
\usepackage{booktabs}
\usepackage{newfloat}
\usepackage{listings}
\usepackage{xcolor}
\definecolor{codegreen}{rgb}{0,0.6,0}
\definecolor{codegray}{rgb}{0.5,0.5,0.5}
\definecolor{codepurple}{rgb}{0.58,0,0.82}
\definecolor{backcolour}{rgb}{0.95,0.95,0.92}

\DeclareCaptionStyle{ruled}{labelfont=normalfont,labelsep=colon,strut=off} 
\floatstyle{ruled}
\newfloat{listing}{tb}{lst}{}
\floatname{listing}{Listing}
\usepackage{hyperref} 
\nocopyright 

\title{Planning in the Dark: LLM-Symbolic Planning Pipeline without Experts}
\author {
    Sukai Huang,
    Nir Lipovetzky and 
    Trevor Cohn\thanks{Now at Google DeepMind}
}
\affiliations {
    School of Computing and Information Systems, The University of Melbourne, Australia\\
    sukaih@student.unimelb.edu.au, \{nir.lipovetzky,
    trevor.cohn\}@unimelb.edu.au \\
}

\usepackage{bibentry}

\begin{document}

\maketitle 

\begin{abstract}
    Large Language Models (LLMs) have shown promise in solving natural language-described planning tasks, but their direct use often leads to inconsistent reasoning and hallucination. While hybrid LLM-symbolic planning pipelines have emerged as a more robust alternative, they typically require extensive expert intervention to refine and validate generated action schemas. It not only limits scalability but also introduces a potential for biased interpretation, as a single expert's interpretation of ambiguous natural language descriptions might not align with the user's actual intent. To address this, we propose a novel approach that constructs an action schema library to generate multiple candidates, accounting for the diverse possible interpretations of natural language descriptions. We further introduce a semantic validation and ranking module that automatically filter and rank the generated schemas and plans without expert-in-the-loop. The experiments showed our pipeline maintains superiority in planning over the direct LLM planning approach. These findings demonstrate the feasibility of a fully automated end-to-end LLM-symbolic planner that requires no expert intervention, opening up the possibility for a broader audience to engage with AI planning with less prerequisite of domain expertise\footnote{The code is available at \url{https://anonymous.4open.science/r/Official-LLM-Symbolic-Planning-without-Experts-7466}}.
\end{abstract}

%
\section{Introduction}

\begin{figure}[t]
    \centering
    \includegraphics[width=0.9\columnwidth]{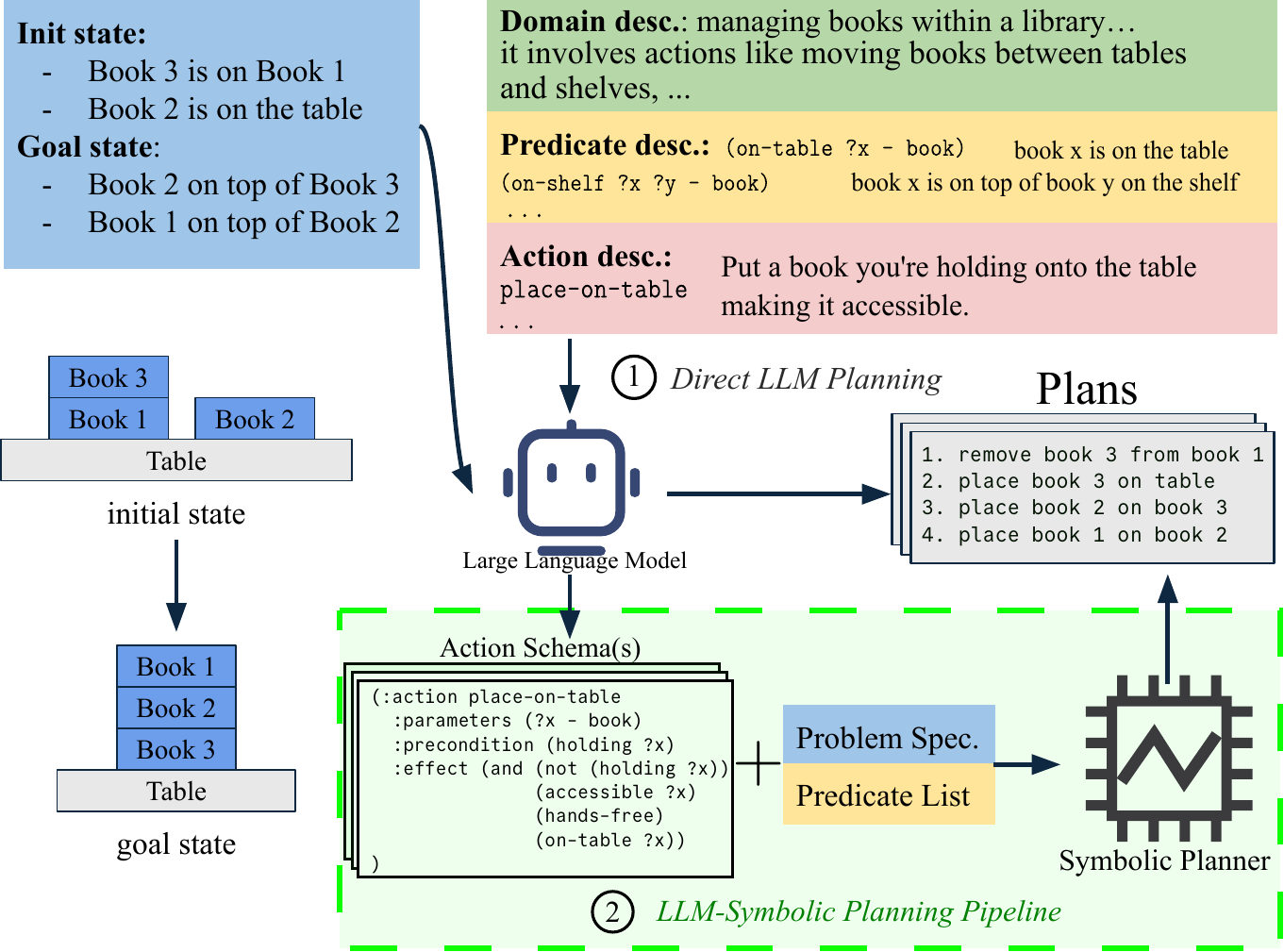}
    \caption{An overview of direct plan generation vs. LLM-symbolic planning pipelines.}
    \label{fig:overview_llm_planning}
\end{figure}

The advent of Large Language Models (LLMs) has opened new avenues for solving natural language-described planning tasks \citep{kojima2022large}. However, direct plan generation using LLMs, while seemingly straightforward, has been criticized for inconsistent reasoning and hallucination, which undermines their reliability in critical planning scenarios \citep{Valmeekam2022PlanBenchAE, valmeekam2023planning, Huang2023ASO}. In response, researchers have advocated for more robust approaches that combine the flexibility of LLMs with the correctness of symbolic planning to solve planning tasks \citep{Pallagani2024OnTP, oswald2024large}. 
To improve the soundness of generated plans, a hybrid LLM-symbolic planning pipeline has emerged. As shown in Figure~\ref{fig:overview_llm_planning}, instead of relying solely on LLMs to generate sequences of action plans through in-context learning, this pipeline begins by leveraging LLMs to extract abstract symbolic action specifications from natural language descriptions, known as \emph{action schemas}. These schemas define the essential components of an action in a structured format understandable by symbolic planners. Once these schemas are generated, a classical planner can take over to search for feasible plans that fulfill the task specifications \citep{Liu2023LLMPEL, Silver2023GeneralizedPI, guan2023leveraging, kambhampatiposition2024}.

\begin{figure}[t]
    \centering
    \includegraphics[width=0.99\columnwidth]{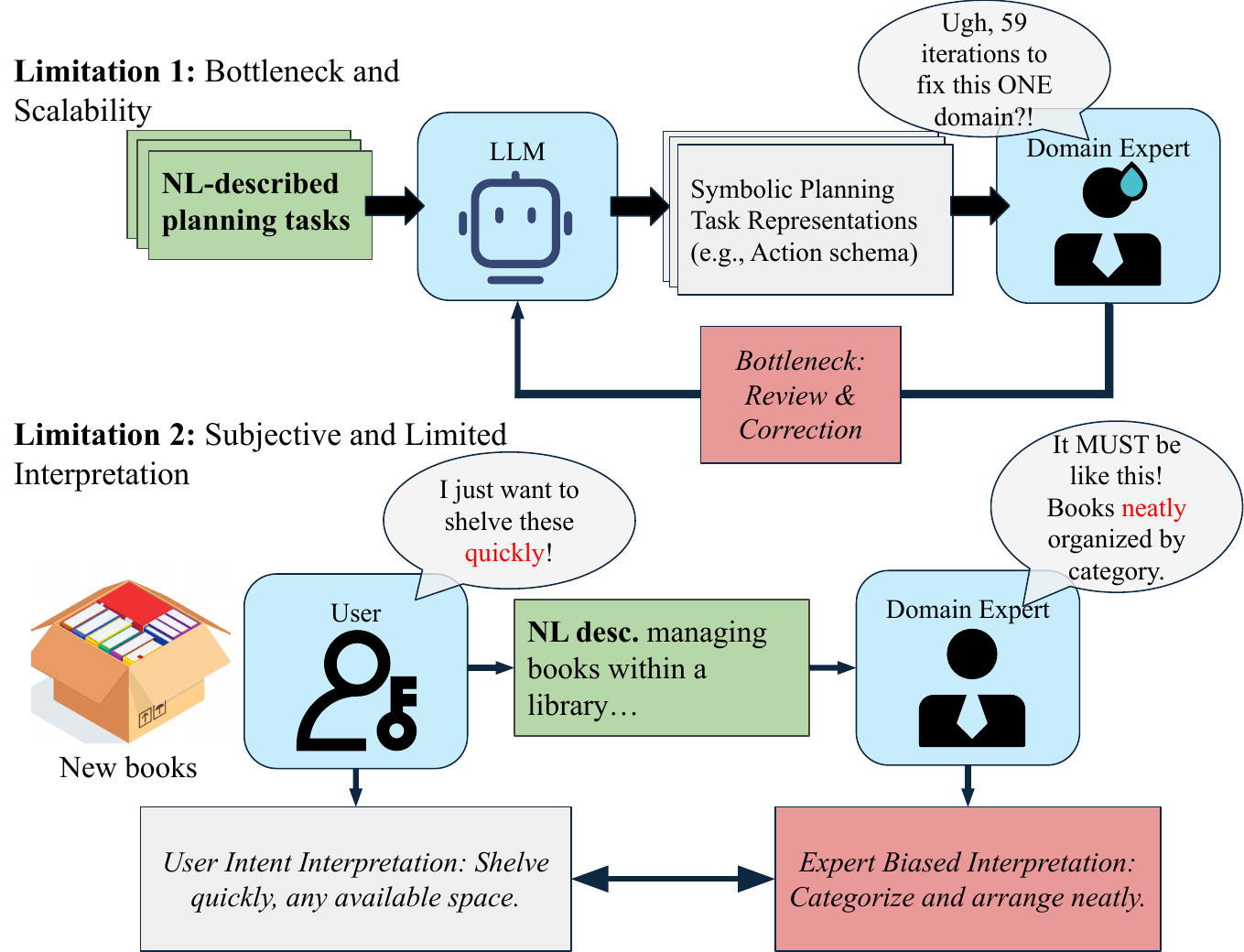}
    \caption{Illustration of the two limitations of expert-dependent LLM-symbolic planning pipelines}
    \label{fig:limitation_illu}    
\end{figure}

Yet, this method is brittle, as a single missing or contradictory predicate in an action schema can prevent the planner from finding a valid plan.
Thus, current pipelines often require multiple iterations of expert intervention to refine and validate the generated action schemas. For instance, \citet{guan2023leveraging} reported that the expert took 59 iterations to fix schema errors for a single task domain. This process demands substantial time and expertise, which significantly hinders the \emph{\textbf{scalability}} of the method. 
More critically, due to budget constraints, often only one expert is involved in the process. This creates a critical vulnerability: the potential for interpretation mismatch between the expert and the user. Experts, while knowledgeable, inevitably bring their own \emph{\textbf{subjective interpretations}} to the task descriptions, often formalizing them in a single, specific way. This limits the system to a \emph{\textbf{single perspective}} of the task. However, unlike formal language designed to have an exact, context-independent meaning, natural language inherently contains ambiguities that yield \emph{diverse} valid interpretations of the same description. This ambiguity suggests that a straightforward, one-to-one mapping from natural to formal languages -- a typical case when relying on a single expert -- risks overlooking the interpretation that the user actually intended \citep{moravcsik1983natural} (see Figure~\ref{fig:limitation_illu}).

Regarding the issue with reliance on expert intervention, we propose a novel pipeline that eliminates this dependency. Specifically, our approach introduces two key innovations:
\begin{itemize}
    \item We construct an \emph{action schema library} to generate multiple candidates, a strategy that has been overlooked in prior work despite being a natural fit for capturing the inherent ambiguity in natural language. By leveraging this library, we also increase the likelihood of obtaining \textbf{\emph{solvable}} action schema sets -- those have at least one valid plan that can be found by a planner.   
    \item We leverages sentence encoders\footnote{Sentence encoders are neural network models that transform sentences into dense vector representations, capturing semantic meaning} to automatically \emph{validate and filter generated action schemas}. This module ensures that the generated schemas closely align with the task descriptions in the semantic space, effectively acting like expert feedback.
\end{itemize}
Our experiments demonstrate that without expert intervention, our pipeline generates sound action plans competitive with direct LLM-based plan generation, even in short-horizon planning tasks. Importantly, our approach offers multiple schema sets and plan candidates, preserving the diversity of interpretations inherent in ambiguous natural language descriptions.

\section{Related Work}

\begin{figure*}[t]
        \centering
        \includegraphics[width=0.99\textwidth]{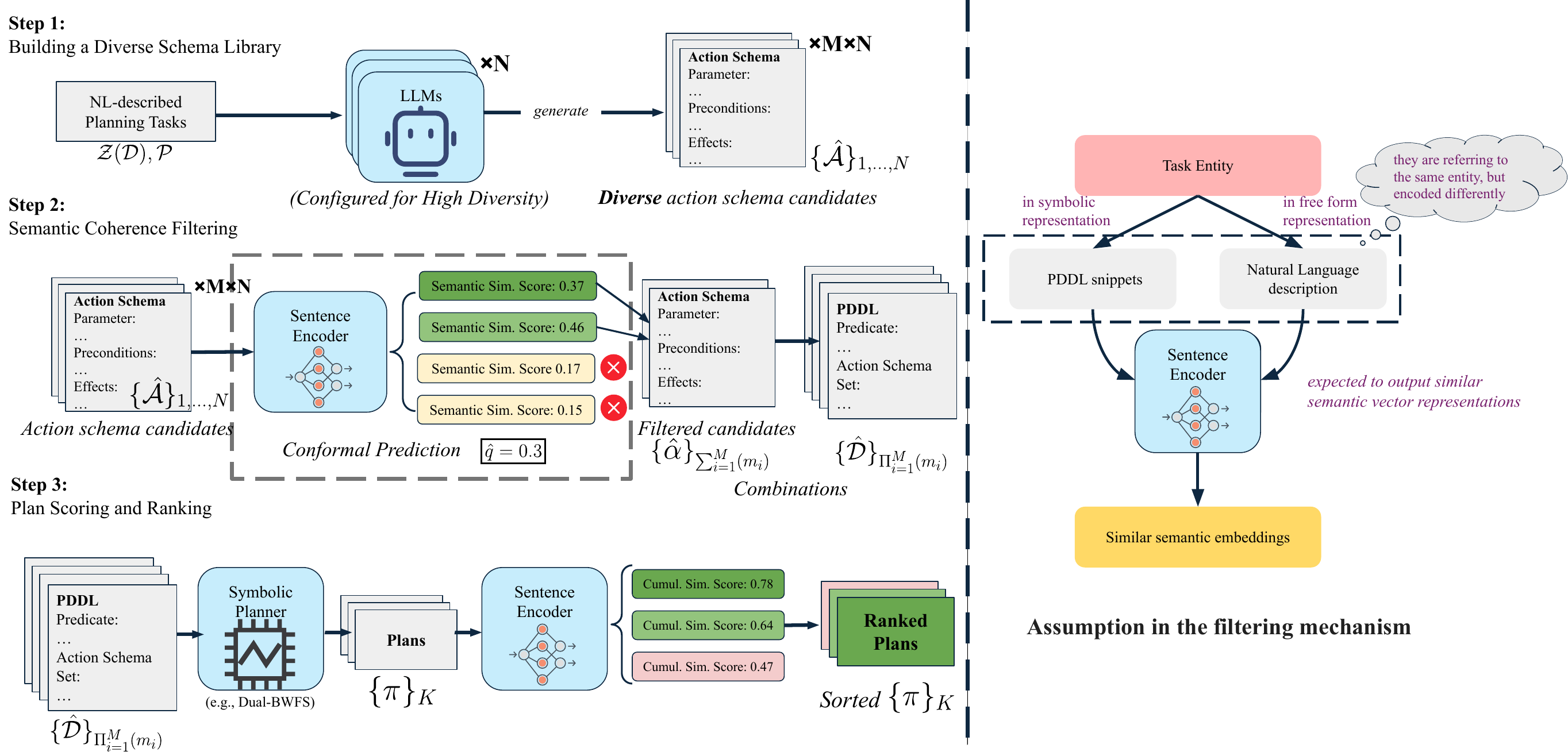}
        \caption{An overview of the proposed pipeline, it first constructs diverse action schema candidates to cover various interpretations of the natural language descriptions. Then, it filters out low-confidence candidates to ensure the generation candidates are semantically aligned with the descriptions. Lastly, it produces and ranks multiple plans using a symbolic planner. The filtering mechanism is grounded in the concept of semantic equivalence across different representations of the same content.}
        \label{fig:overview_our_pipeline}

\end{figure*}

\textbf{Direct Plan Generation with LLMs:}\quad The use of LLMs for direct action plan generation has been explored across various domains, including embodied tasks \citep{Wang2023DescribeEP, xiang2024language}, and other language grounding environments \citep{Ahn2022DoAI, huang2022language}. 
These approaches are built upon the idea that LLMs' reasoning capabilities can be effectively elicited through in-context learning techniques, particularly the Chain-of-Thought (CoT) approach. CoT prompts the model to generate a series of intermediate reasoning steps before arriving at the final answer, resulting in more coherent and logically sound reasoning \citep{wei2022chain}. Building upon CoT, \citet{yao2024tree} proposed Tree-of-Thought (ToT) framework, which explores multiple reasoning pathways, generating diverse plans and ranking them based on self-verification heuristics. These heuristics are verbalized confidence scores produced by LLMs themselves, a method supported by studies showing that LLMs are effective as zero-shot ranking models \citep{lin2022teaching, hou2023large, zhuang2023open}.


\noindent\textbf{Criticism and Hybrid Planning:}\quad Despite the promising results, researchers have raised concerns about the reliability and soundness of LLM-generated plans \citep{Valmeekam2022PlanBenchAE, valmeekam2023planning, Huang2023ASO}. A critical issue highlighted by \citet{kambhampatiposition2024} is that planning and reasoning tasks are typically associated with System 2 competency, which involves slow, deliberate, and conscious thinking \citep{Sloman1996TheEC, kahneman2011thinking}. However, LLMs, being essentially text generators, exhibit constant response times regardless of the complexity of the question posed. This behavior suggests that no first-principle reasoning is occurring, contradicting the expectations for true planning capabilities. To this end, researchers have explored hybrid approaches that utilize LLMs to generate symbolic representations of tasks, which are then processed by external symbolic planners to search for feasible plans \citep{Liu2023LLMPEL, guan2023leveraging}. However, existing pipelines emphasize the necessity of expert intervention for action schema validation and refinement. While \citet{kambhampatiposition2024} proposed using LLMs as semi-expert critics to assess output quality, this approach still necessitates expert involvement for final decision-making. In contrast, our work strives to reduce the dependency on expert intervention, offering a more accessible approach to hybrid LLM-symbolic planning that also addresses the inherent ambiguity in natural language descriptions.       

\noindent\textbf{Acquiring Action Models:}\quad It is important to acknowledge the rich body of work that has focused on acquiring symbolic action models before the advent of LLMs \citep{mourao2008using,aineto2019learning}. Our approach introduces significant complexity due to the inherent ambiguity in natural language descriptions. However, it also offers greater accessibility, allowing users to describe domains without requiring specialized knowledge of planning formalisms.

\section{Problem Setting and Background}
We consider a scenario where an agent generates action plans for natural language-described planning tasks. A task description typically consists of: (1) a domain description outlining general task information and possible high-level actions, and (2) a problem instance description specifying the initial and goal states. The study of LLM-symbolic planning pipelines is grounded in the formal framework of classical planning, which relies on symbolic representations of planning tasks. These representations are typically expressed using the Planning Domain Definition Language (PDDL) \citep{McDermott1998PDDLthePD, haslum2019introduction}. In brief, a PDDL description is defined by $\langle \mathcal D, \Pi_{\mathcal D}\rangle$, where:
\begin{itemize}
    \item $\mathcal D = \langle \mathcal P, \mathcal A \rangle$ is the domain specification: $\mathcal P$ is the set of predicates that can either hold true or false, and $\mathcal A$ is the set of action schemas. Each action schema $\alpha \in \mathcal A$ is defined as a tuple $\alpha = \langle \mathit{par}, \mathit{pre}, \mathit{eff} \rangle$, where $\mathit{par}$ details the parameters, and $\mathit{pre}$ and $\mathit{eff}$ are the preconditions and effects, respectively. Both $\mathit{pre}$ and $\mathit{eff}$ are typically expressed as conjunctive logical expressions using predicate logic.
    \item $\Pi_{\mathcal D} = \langle \mathcal O, \mathcal I, \mathcal G \rangle$ is the problem instance: $\mathcal O$ is the set of objects to interact with, $\mathcal I$ is the initial state, and $\mathcal G$ is the goal state that the agent needs to achieve.
\end{itemize}
\noindent A solution to the planning task is a sequence of grounded actions $\left(\pi = (a_0, ..., a_n)\right)$ that transforms the initial state $\mathcal I$ to the goal state $\mathcal G$. Each grounded action $a_i$ is an instantiation of an action schema $\alpha \in \mathcal A$ and predicates, where the parameters in $\alpha$ are replaced with specific objects from $\mathcal O$. 

To bridge natural language descriptions and formal planning representations, we introduce a natural language proxy layer, denoted as $\mathcal Z(\cdot)$, for these task specifications. For example, $\mathcal Z(\mathcal D)$ represents the natural language equivalent of the domain specification $\mathcal D$. The two approaches, \emph{direct LLM planning} and \emph{LLM-symbolic planning}, can then be expressed in Eq~1 and Eq~2, respectively:

\begin{flalign}
    \pi &\sim P_\text{LLM}(\cdot \mid \mathcal Z(\mathcal D), \mathcal Z(\Pi_{\mathcal D})) \\
    \hat{\mathcal A} \sim P_\text{LLM} &\left(\cdot \mid \mathcal Z(\mathcal D)\right) ; \Pi_{\mathcal D} \sim P_\text{LLM}\left(\cdot \mid \mathcal Z(\Pi_{\mathcal D}) \right)  \nonumber\\
    &\pi = f\left(\langle \hat{\mathcal A}, \mathcal P \rangle, \Pi_{\mathcal D}\right)
\end{flalign}
\noindent In these equations, $P_\text{LLM}(\cdot)$ represents the generation process of LLMs, and $f$ is the symbolic planner that search for sound plans. While we largely adhere to the problem setting of previous research (e.g., \citet{Liu2023LLMPEL}, \citet{guan2023leveraging}), we introduce a crucial refinement by specifying a precise predicate set ($\mathcal{P}$) for each domain descriptions. This controlled setting addresses a key challenge in evaluating across different methodologies. Without a standardized predicate set, variations in domain understanding can lead to diverse and potentially incomparable outputs, hindering meaningful evaluation.

\section{Methodology}

As illustrated in Figure~\ref{fig:overview_our_pipeline}, the proposed pipeline stands in contrast to existing expert-dependent approaches and consists of three key steps: \emph{(1) Building a Diverse Schema Library (\S \ref{sec:method_pool}), (2) Semantic Coherence Filtering (\S~\ref{sec:method_semantic}) and (3) Plan Scoring and Ranking (\S~\ref{sec:method_ranking})}. 

\subsection{Building a Diverse Schema Library}
\label{sec:method_pool}
A key challenge in translating natural language descriptions into symbolic action schemas is the inherent ambiguity of language itself. Different interpretations of the same description can lead to variations in action schemas, impacting the downstream plan generation process. To ensure we explore a wide range of interpretations and effectively cover the user's intent, we utilize multiple LLM instances, denoted as $\{P_{\text{LLM}}^1, P_{\text{LLM}}^2, ..., P_{\text{LLM}}^N\}$, and set their temperature hyperparameter high to encourage diverse outputs. Each will then generate its own set of action schemas $\hat{\mathcal A}_i \sim P_{\text{LLM}}^i(\cdot \mid \mathcal Z(\mathcal D))$, where $\hat{\mathcal A}_i =  (\hat{\alpha}_{i1}, \hat{\alpha}_{i2}, ..., \hat{\alpha}_{iM})$. Here, $\hat{\alpha}_{ij}$, where $i\in [1,..N]$ and $j\in [1,...,M]$, represents the generated action schema of $j$-th action in the domain by the $i$-th LLM instance. 

The generated schemas $\hat{\alpha}_{ij}$ from all models are then aggregated into a single library. Since each domain comprises $M$ actions, a ``set'' of action schemas refers to a complete collection where each action in the domain is associated with one corresponding schema.
Therefore, all possible combination of action schemas within the library can generate approximately $\binom{N}{1}^M$ different sets of action schemas.

In addition, existing pipelines rely heavily on expert intervention, partly because individual LLMs struggle to generate \emph{solvable} sets of schemas -- those that a planner can successfully use to construct a plan. This reliance becomes even more pronounced as the number of actions increases, with the probability of obtaining a solvable set of schemas from a single LLM diminishing exponentially. In contrast, our approach, by constructing a diverse pool of action schema sets, substantially improves the probability of finding a solvable set. Our analysis (detailed in Appendix~\ref{app:pool_analysis}) demonstrates that, under reasonable assumptions, this probability can increase from less than 0.0001\% with a single LLM to over 95\% when using multiple LLM instances.

Note that the \emph{solvability} of a set of action schemas can be efficiently verified by leveraging the \emph{completeness} feature of modern symbolic planners (see Appendix~\ref{app:symbolic_planner_details}). If a plan can be found for a given problem using the generated schemas, the set is deemed solvable. Building an action schema library thus offers two key advantages: (1) the use of multiple LLM instances enables the generation of diverse candidate sets, allowing for a more comprehensive exploration of possible interpretations of ambiguous natural language descriptions, and (2) it significantly increases the likelihood of obtaining solvable schema sets without relying on expert intervention.

\subsection{Semantic Coherence Filtering}
\label{sec:method_semantic}




The previous method alone faces two limitations. First, as task complexity grows, the ``brute-force'' approach of combining and evaluating all possible sets becomes increasingly inefficient. Second, solvability does not guarantee semantic correctness -- schemas may not accurately reflect the task descriptions, potentially leading to incorrect or nonsensical plans. Therefore, it is crucial to implement a filtering mechanism that autonomously assesses the semantic correctness of individual action schemas, filtering out low-quality candidates before they enter the combination process. 

Our approach is grounded in the concept of semantic equivalence across different representations of the same content, as discussed by \citet{weaver1952translation} in his memorandum ``Translation.'' Weaver emphasized that the most effective way to translate between languages is to go deeper to uncover a shared ``common base of meaning'' between language representations, illustrating this by noting that ``a Russian text is really written in English, but it has been encoded using different symbols.'' This principle is crucial in our context, where task descriptions in natural language and their corresponding structured symbolic representations should exhibit high semantic similarity, reflecting the same shared meaning despite different syntactic forms (see right side of Figure~\ref{fig:overview_our_pipeline}).\todo[inline, color=green]{I try to rewrite the Weaver's part to make it align with the context of our work.}

Recent developments in language models as code assistants \citep{chen2021evaluating, roziere2023code} further support this assumption, demonstrating that these models can decode the underlying semantics of structured symbolic representations. Inspired by this, we propose a filtering step that leverages a sentence encoder $E(\cdot)$ to generate embeddings for both the action descriptions $E(\mathcal Z(\alpha))$ and the generated schemas $E(\hat{\alpha})$. Then, we compute the cosine similarity between these embeddings to quantify semantic relatedness and filter out action schemas with low scores.

Specifically, we employ a conformal prediction (CP) framework (see Appendix~\ref{app_sec:cp_algo}) to statistically guarantee that true positive action schema candidates have a high probability of being preserved while minimizing the size of the filtered set \citep{sadinle2019least}. In this process, a threshold $\hat{q}$ will be calculated based on a user-specified confidence level $1-\epsilon$. Action schemas with cosine similarity scores below this threshold are filtered out from the library.

This process (illustrated in \emph{step 2} of Figure~\ref{fig:overview_our_pipeline}) significantly reduces the \emph{number of candidate sets of action schemas} to $\Pi_{i=1}^{M}(m_i)$, where $m_i$ is the number of action schemas that pass the semantic validation for the $i$-th action. This pre-filtering approach not only reduces the computational load on the symbolic planner, increasing efficiency, but also ensures that generated schemas closely align with the semantic meaning of the task descriptions.

\subsection{Finetuning with Manipulated Action Schemas}
Hard negative samples have been shown to enhance representation learning by capturing nuanced semantic distinctions \citep{robinsoncontrastive2023}. In our context, we found that structured action schemas are particularly ideal for generating hard negatives. By manipulating predicates in the precondition or effect expressions of true action schemas, we create hard negatives with subtle differences. During finetuning, a triplet loss function is employed, where each training sample consists of a triplet: the natural language description of an action $(\mathcal Z(\alpha))$, the true action schema $(\alpha)$, and a negative sample $(\alpha^\text{neg})$. A negative sample is of three types -- \emph{(1) Easy Negatives}: action schemas from other planning domains (inter-domain mismatch); \emph{(2) Semi-Hard Negatives}: action schemas from the same domain but referring to different actions (inter-domain mismatch); and \emph{(3) Hard Negatives}: manipulated versions of the true action schema in the same domain (see Table~\ref{tab:hard_negative_manipulation}). 

\begin{figure}[t]
    \centering
    \includegraphics[width=0.73\columnwidth]{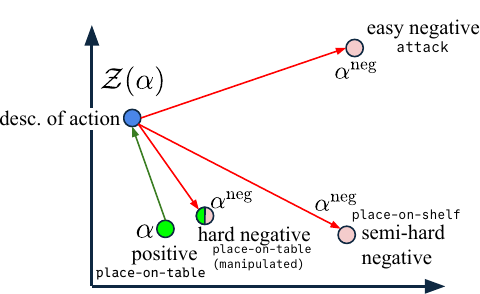}
    \caption{Finetuning sentence encoder with triplet loss}
    \label{fig:training_embedding_triplet_illu}
\end{figure}

Through this process, the sentence encoder learns to embed natural language descriptions closer to their corresponding action schemas while distancing them from negative samples in the semantic space (see Figure~\ref{fig:training_embedding_triplet_illu}).

\begin{table*}[t]
    \centering
    \caption{Types of Manipulations for Generating Synthesized Hard Negative Action Schemas in Training Data}
    \label{tab:hard_negative_manipulation}
    \resizebox{0.80\textwidth}{!}{%
    \begin{tabular}{@{}cll@{}}
    \toprule
    Manipulation Type & Description                                                                                           & Example                                                                                                                                                                                                       \\ \midrule
    Swap              & \begin{tabular}[c]{@{}l@{}}Exchanges a predicate between\\ preconditions and effects\end{tabular}     & \begin{tabular}[c]{@{}l@{}}Precondition: $\texttt{(at ?x ?y)}$\\ Effect: $\texttt{(not (at ?x ?z))}$\\ $\rightarrow$\\ Precondition: $\texttt{(not (at ?x ?z))}$\\ Effect: $\texttt{(at ?x ?y)}$\end{tabular} \\ \midrule
    Negation          & \begin{tabular}[c]{@{}l@{}}Negates a predicate in either\\ preconditions or effects\end{tabular}      & \begin{tabular}[c]{@{}l@{}}Precondition: $\texttt{(clear ?x)}$\\ $\rightarrow$\\ Precondition: $\texttt{(not (clear ?x))}$\end{tabular}                                                                       \\ \midrule
    Removal           & \begin{tabular}[c]{@{}l@{}}Removes a predicate from \\ either preconditions or effects\end{tabular}   & \begin{tabular}[c]{@{}l@{}}Precondition: $\texttt{(and (on ?x ?y) (clear ?x))}$\\ $\rightarrow$\\ Precondition: $\texttt{(on ?x ?y)}$\end{tabular}                                                            \\ \midrule
    Addition          & \begin{tabular}[c]{@{}l@{}}Adds mutually exclusive (mutex)\\predicates to preconditions or \\effects  \citep{helmert2009concise}\tablefootnote{Mutexes are predicates that cannot be true simultaneously, e.g., one cannot hold a book and have it on a table simultaneously.}\end{tabular} & \begin{tabular}[c]{@{}l@{}}Effect: $\texttt{(on-table ?x)}$\\ $\rightarrow$\\ Effect: $\texttt{(and (on-table ?x) (holding ?x))}$\end{tabular}                                                                     \\ \bottomrule
    \end{tabular}%
    }
    \end{table*}

\subsection{Plan Generation and Ranking}
\label{sec:method_ranking}
Action schemas that more accurately represent the intended tasks described in natural language are likely to yield higher-quality, more reliable plans. Leveraging this causal relationship, we assess and rank the generated plans based on the cumulative semantic similarity scores of their constituent action schemas. Specifically, we feed each solvable set of action schemas into a classical planner, which generates a corresponding plan. Then, the ranking score for a plan is calculated as $\sum_{i=1}^M \frac{E(\mathcal Z( \alpha_i)) \cdot E( \hat{\alpha_i})}{\|E(\mathcal Z( \alpha_i))\| \|E( \hat{\alpha_i})\|}$, where $\mathcal Z( \alpha_i)$ is the natural language description of the $i$-th action in the domain and $\hat{\alpha_i}$ is the corresponding generated action schema. It ensures that the structured symbolic model comprising the plans are semantically aligned with the descriptions of the planning domain (see \emph{step 3} in Figure~\ref{fig:overview_our_pipeline}). Furthermore, this approach allows for optional \textbf{lightweight expert intervention} as a final, non-iterative step. By presenting the ranked schema sets and their corresponding plans, experts can determine the most appropriate one, providing a balance between autonomy and expert guidance.\todo[inline, color=green]{I rewrite a bit to make it more clear.}

Overall, our pipeline bridges the gap between ambiguous task descriptions and the precise requirements of symbolic planners. By generating a diverse pool of action schemas and leveraging semantic similarity for validation and ranking, we achieve two key advancements. First, we reduce the dependency on expert intervention, making the process more accessible and efficient. Second, we preserve the inherent ambiguity of natural language, offering users multiple valid interpretations of the task and their corresponding plans.

\section{Experiments}
Our experiments test the following hypotheses: \textbf{(H1)} Semantic equivalence across different representations, as discussed by Weaver, holds true in our context. \textbf{(H2)} Ambiguity in natural language descriptions leads to multiple interpretations. \textbf{(H3)} Our pipeline produces multiple solvable candidate sets of action schemas and plans without expert intervention, providing users with a range of options. \textbf{(H4)} Our pipeline outperforms direct LLM planning approaches in plan quality, demonstrating the advantage of integrating LLM with symbolic planning method.

\begin{figure}[t]
    \centering
    \includegraphics[width=\columnwidth]{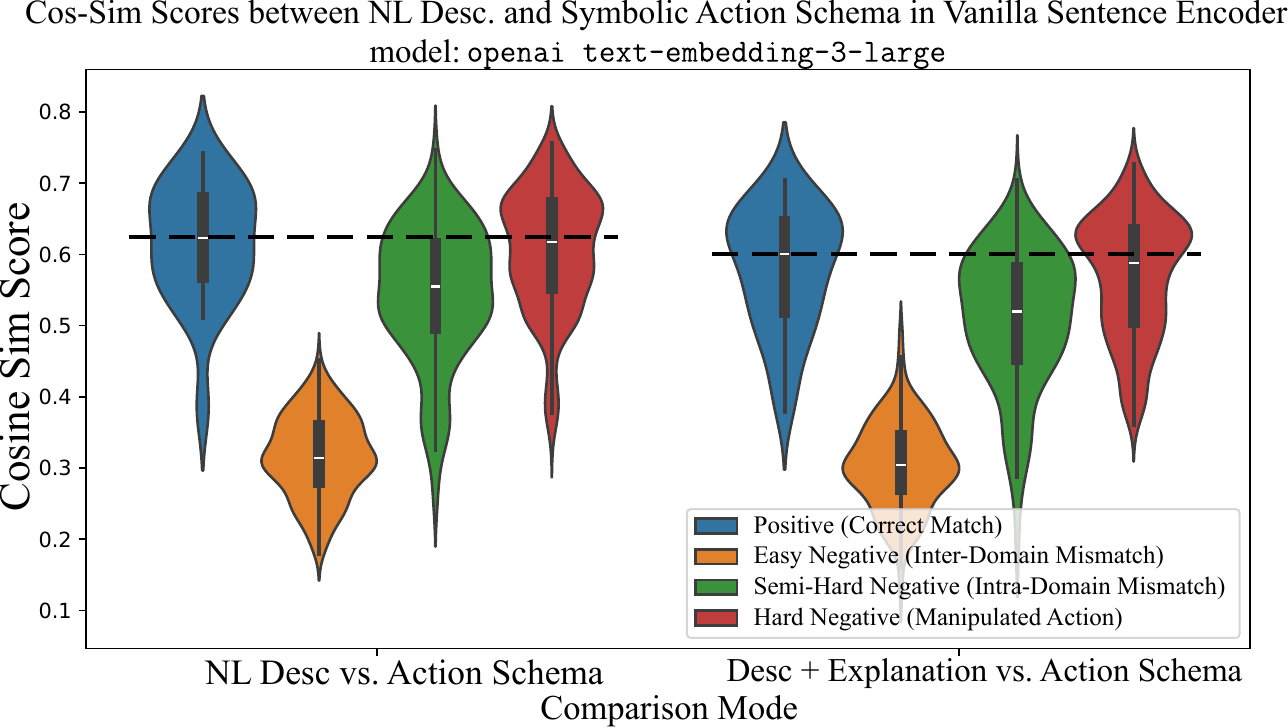}
    \caption{The pre-trained sentence encoder demonstrates semantic alignment for matched action schemas and misalignment for mismatched ones, supporting \textbf{H1}.}
    \label{fig:cosine_sim_comparison_before_finetune}

\end{figure}

\begin{figure*}[t]
    \centering
    \begin{minipage}{0.36\textwidth}
        \centering
        \includegraphics[width=0.70\textwidth]{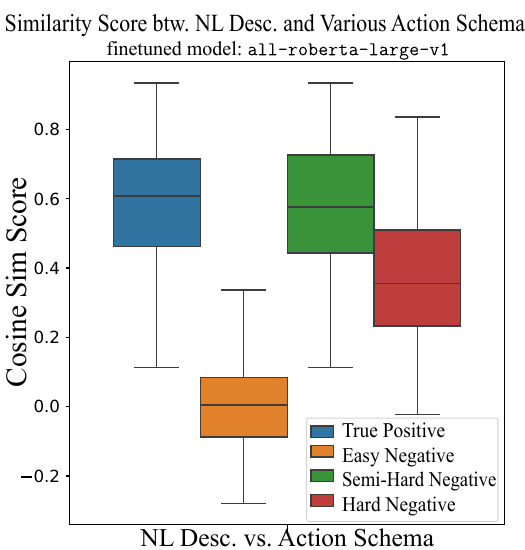}
        \caption{The sentence encoder enhances the identification of mismatched pairs by fine-tuning with negative samples.}
        \label{fig:cosine_sim_comparison_after_finetune}
    \end{minipage}
    \hspace{0.2cm}
    \begin{minipage}{0.256\textwidth}
        \centering
        \includegraphics[width=0.93\textwidth]{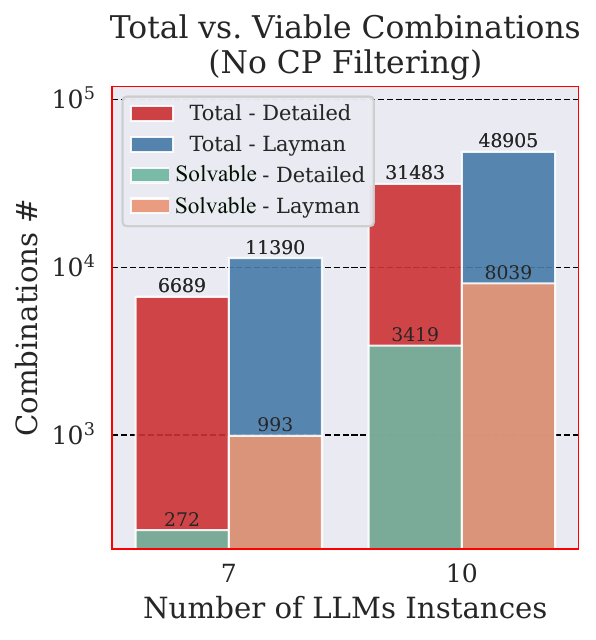}
        \caption{More LLM instances generate more solvable schemas without an expert.}
        \label{fig:total_vs_viable_combinations_cp_False}
    \end{minipage}
    \hspace{0.2cm}
    \begin{minipage}{0.326\textwidth}
        \centering
        \includegraphics[width=\textwidth]{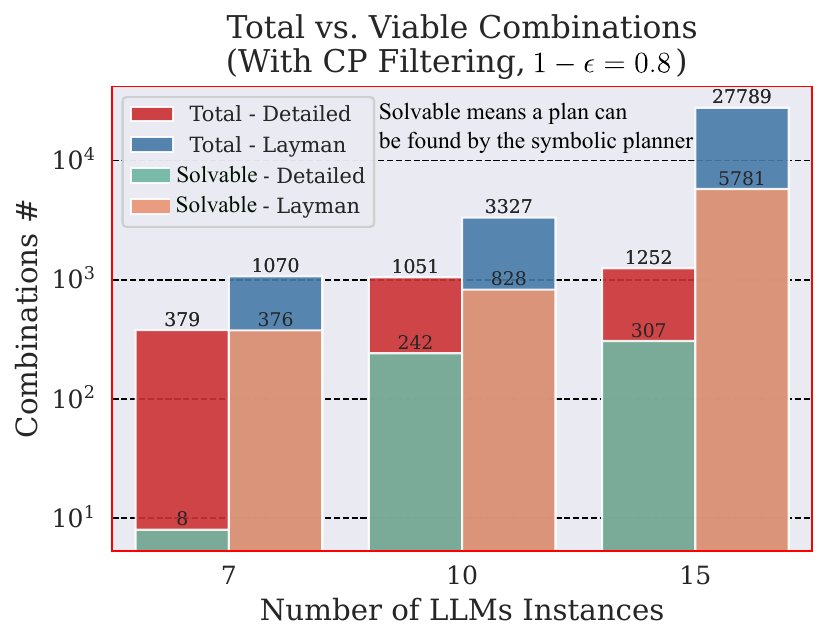}
        \caption{With CP, a large number of candidates are pruned, thereby improving efficiency.}
        \label{fig:total_vs_viable_combinations_cp_True}
    \end{minipage}
\end{figure*}

\subsection{Experimental Setup}
\noindent \textbf{Task and Model Setup.}\quad We introduces several key enhancements that distinguish it from previous work. \emph{(1) Novel Test Domains:} We carefully selected three test domains ensuring they are unfamiliar to LLMs -- \textbf{Libraryworld}: a modified version of the classic Blockworld domain; \textbf{Minecraft}: resource gathering and crafting domain inspired by the game Minecraft; and \textbf{Dungeon}: a domain originally proposed by \citet{chrpa2017fifth}. This approach addresses a significant issue: many IPC\footnote{International Planning Competition, a benchmark event for automated planning systems using PDDL.} domains have likely been leaked into LLM training data (see Appendix~\ref{app:leak_investigation_methodology}). For training and calibration of the sentence encoder, we used domains from IPC and PDDLGym \citep{silver2020pddlgym}. \emph{(2) LLM Selection:} We use the open-source GLM \citep{hou2024chatglm} over proprietary models like GPT-4, aligning with our commitment to accessible planning systems. \emph{(3) Ambiguity Examination:} We tested our pipeline on two types of task descriptions to assess the impact of ambiguity -- (a) \textbf{\emph{detailed}} descriptions following the established style of \citet{guan2023leveraging}, and (b) \textbf{\emph{layman}} descriptions provided by five non-expert participants\footnote{Business school students with no prior knowledge of PDDL programming or computational logic} who, unfamiliar with PDDL, described the domains and actions based on reference PDDL snippets. \emph{(4) Symbolic Planner:} We used \textit{DUAL-BWFS} \citep{Lipovetzky2017BestFirstWS} planner for plan generation as well as checking if the generated schema sets are solvable. \emph{(5) LLM Prompt Engineering:} We use the CO-STAR and CoT framework to guide LLMs in generating outputs (see Appendix~\ref{app:experimental_setup}). 

\noindent \textbf{Baselines.}\quad We evaluate our pipeline against two key baselines: \emph{(1) The previous LLM-symbolic planning pipeline} proposed by \citet{guan2023leveraging}, which involves expert intervention for action schema validation and refinement; and \emph{(2) Direct LLM-based planning} using Tree-of-Thought (ToT) \citep{yao2024tree}, which generates multiple plans and ranks them based on self-verification heuristics.

\subsection{Semantic Equivalence Analysis}



To investigate \textbf{H1}, we initially assessed the cosine similarity of sentence embeddings for matched and mismatched pairs of action schemas and natural language descriptions. We used two large-scale sentence encoders, \textit{text-embedding-3-large} and \textit{sentence-t5-xl}, and both models demonstrated higher cosine similarity between matched pairs compared to mismatched ones (see Figure~\ref{fig:cosine_sim_comparison_before_finetune}). This confirmed the presence of semantic equivalence across representations. For practical implementation, we finetuned a lightweight encoder \textit{all-roberta-large-v1} to further enhance discrimination. As shown in Figure~\ref{fig:cosine_sim_comparison_after_finetune}, while the model's performance on semi-hard negatives showed limited improvement, finetuning significantly boosted its ability to identify \emph{hard negatives} -- mismatched pairs with subtle differences. This targeted enhancement is particularly valuable because these hard negatives closely resemble common LLM errors in action schema generation.

\subsection{Pipeline Performance and Efficiency}
\begin{table}[h]
    \centering
    \caption{Contrasts Our Pipeline with Existing Works}
    \label{tab:contrast_with_baselines}
    \resizebox{0.99\columnwidth}{!}{%
    \begin{tabular}{@{}lcccccc@{}}
    \toprule
    Model         & \begin{tabular}[c]{@{}c@{}}Planning\\ Mechanism\end{tabular} & \begin{tabular}[c]{@{}c@{}}Expert\\ Input \#\end{tabular} & \begin{tabular}[c]{@{}c@{}}Action\\ Schema \#\end{tabular} & Plan \#  & \begin{tabular}[c]{@{}c@{}}Heuristic\\ Type\end{tabular} & \begin{tabular}[c]{@{}c@{}}Soundness\\ Guarantee\end{tabular} \\ \midrule
    \begin{tabular}[c]{@{}l@{}}Tree-of-Thought\\ \citep{yao2024tree}\end{tabular} & Pure LLM            & 0                      & N/A              & Multiple & \begin{tabular}[c]{@{}c@{}}Self\\ Verification\end{tabular} & No                  \\ \midrule
    \citet{guan2023leveraging} & Hybrid & $\approx 59$           & Single & Single   & \begin{tabular}[c]{@{}c@{}}Expert\\ Validation\end{tabular}            & Yes                 \\ \midrule
    \textbf{Ours}          & Hybrid & $\leq 1$               & Multiple         & Multiple & \begin{tabular}[c]{@{}c@{}}Semantic\\ Sim. Scores\end{tabular}   & Yes                 \\ \bottomrule
    \end{tabular}%
    }
\end{table}

Our pipeline's performance and efficiency are highlighted through several key observations. Firstly, the use of action schema library effectively produces \emph{solvable} action schema sets without requiring expert-in-the-loop, as demonstrated in Figure~\ref{fig:total_vs_viable_combinations_cp_False}. Notably, deploying 10 LLM instances is sufficient to generate solvable schema sets for all test domains, supporting \textbf{H3}. Secondly, Figures \ref{fig:total_vs_viable_combinations_cp_False} and \ref{fig:total_vs_viable_combinations_cp_True} reveal a clear pattern: when confronted with inherently ambiguous \textbf{\emph{layman}} descriptions from non-expert participants, our pipeline generates a significantly increased number of distinct solvable schema sets (e.g., from 3419 to 8039 when LLM\# = 10 w/o CP)\todo[inline, color=green]{add e.g.}, thereby supporting \textbf{H2}. This phenomenon stems primarily from the varied selection of predicates in the action schemas. Different predicate selections reflect diverse interpretations of the problem, with each set \emph{emphasizing distinct features} deemed critical for planning. For instance, in the \emph{Libraryworld} domain, we observed cases where some schema sets included the `category' of the book when considering actions like stacking books on a shelf. This restriction means that only books of the same category can be stacked together, leading to different plans that reflect various interpretations of the task. The pipeline's ability to generate a range of potential interpretations in response to ambiguous inputs is a critical advantage. It ensures that all intended aspects of the user's description can be captured, even when the description is imprecise or incomplete.

Thirdly, the integration of conformal prediction in the filtering step demonstrates a significant improvement in efficiency, as evidenced by Figure~\ref{fig:total_vs_viable_combinations_cp_True}. With the confidence level $1 - \epsilon$ set to 0.8, the pipeline filtered out a large number of candidates, reducing the total number of combinations to 3.3\% of the original (1051 out of 31483) but meanwhile, the ratio of solvable schemas (verified by the planner) increased from 10.9\% to 23.0\%.\todo[inline, color=green]{add paranthesis verified by the planner} This result strongly supports \textbf{H3}, highlighting the pipeline's ability to efficiently generate solvable and semantically coherent schema sets. See Table~\ref{tab:contrast_with_baselines} for a comprehensive comparison of our pipeline with existing LLM-based planning approaches. Notably, the initial low ratio of solvable schema sets (10.9\%) underscores the challenge faced within the LLM-symbolic planning paradigm, which may explain why expert intervention has been a common practice in the past.

\subsection{Human Evaluation on Plan Quality}
\begin{table}[t]
    \centering
    \caption{Blind plan ranking eval.: Four assessors compared the top two plans from each approach to gold plans.}
    \label{tab:human_eval_4_assessors}
    \resizebox{0.99\columnwidth}{!}{%
    \begin{tabular}{@{}ccccccc@{}}
    \toprule
     & Rank 1st & Rank 2nd & Rank 3rd & Rank 4th & Rank 5th & Avg. Rank \\ \midrule
    \textbf{Gold}      & 14      & 4        & 4        & 1        & 1        & 1.79      \\ 
    \textbf{Ours}      & 4        & 18       & 11       & 5        & 10       & 2.97      \\
    \textbf{ToT}       & 3        & 6        & 12       & 14       & 13       & 3.58      \\ \bottomrule
    \end{tabular}%
    }
\end{table}
To further validate our approach, we conducted a human evaluation comparing the top two plan candidates generated by our pipeline against those from the ToT framework and a gold-standard plan derived from the reference PDDL domain model. Four expert assessors with extensive PDDL experience ranked the plans based on their feasibility in solving the given problems. The results, summarized in Table~\ref{tab:human_eval_4_assessors}, clearly support \textbf{H4}.

For a deeper insight into our pipeline's capabilities, we specifically tested the Sussman Anomaly, a well-known planning problem that requires simultaneous consideration of multiple subgoals, as solving them in the wrong order can undo previous progress (see Figure~\ref{fig:overview_llm_planning}). As shown in Table~\ref{tab:sussman_performance}, ToT approaches using various LLMs, including state-of-the-art models like GPT-4o, consistently fail to solve this problem. The failure arises from the mistaken assumption that the first subgoal mentioned (i.e., placing book 1 on top of book 2) should be addressed first, leading to incorrect plans. Interestingly, GPT-3.5 and GPT-4o exhibited different behaviors when faced with this problem. While GPT-3.5 consistently, yet incorrectly, asserted it had completed the problem, GPT-4o occasionally exhibited awareness of the plan's incompleteness. However, even with this heightened awareness, GPT-4o was unable to identify the correct path within the given depth limit. In contrast, our pipeline generates a range of plans, including suboptimal ones, but excels at identifying and prioritizing the most promising candidates through its ranking process that is based on the cumulative cosine similarity scores of generated action schemas. By strictly adhering to semantic alignment between these schemas and natural language descriptions, and by using a symbolic planner, the system avoids being misled by the tendency -- observed in both humans and LLMs -- to reason in a linear manner. This tendency involves prioritizing subgoals based on their order of appearance rather than considering their underlying logical dependencies. Such linear reasoning can lead to noninterleaved planning, where subgoals are tackled in the order they are presented and each must be fully completed before addressing the next one, which is a pitfall in complex planning problems like the Sussman Anomaly.

.\todo[inline, color=green]{add elaboration on the intuition bias}

\begin{table}[t]
    \centering
    \caption{Performance on Sussman Anomaly problem of ToT approach vs. ours. Both approaches generate multiple plans: ToT uses beam search, while ours generates multiple plans by feeding diverse sets of action schemas into a classical planner, with each set producing its own corresponding plan. Note that different schema sets can result in identical plans.}
    \label{tab:sussman_performance}
    \resizebox{0.99\columnwidth}{!}{%
    \begin{tabular}{@{}lll@{}}
    \toprule
    Model              & Action Plan                                                                                                                                                                                                  & Score             \\ \midrule
    ToT GLM           & \begin{tabular}[c]{@{}l@{}}Take book2 from table, Place book2 on book3,\\ Take book1 from table, Place book1 on book2\end{tabular}                                                                       & Heuristic: 9.0    \\ \midrule
    ToT GPT-3.5 & \begin{tabular}[c]{@{}l@{}}Take book2, Place book2 on book3,\\ Place book1 on table, Place book1 on book2,\\ Remove book3, Place book3 on book2,\\ Place book1 on book2, Check out book1\end{tabular}    & Heuristic: 5.11   \\ \midrule
    ToT GPT-4o (1)    & \begin{tabular}[c]{@{}l@{}}Take book2 from table, Place book2 on book3,\\ Take book1 from table, Place book1 on book2\end{tabular}                                                                       & Heuristic: 8.5    \\ \midrule
    ToT GPT-4o (2)    & \begin{tabular}[c]{@{}l@{}}Take book2 from table, Take book3,\\ Place book3 on table, Place book2 on table,\\ Take book3 from table, Place book2 on table,\\ Place book3 on book1, Take book2 from table,\\ Place book2 on book3, \textbf{Depth limit reached}\end{tabular} & Heuristic: 7.89   \\ \midrule
    Ours GLM (1)     & \begin{tabular}[c]{@{}l@{}}Remove book3 from book1, Take book1 from table,\\ Place book1 on book2, Take book2 from table,\\ Place book2 on book3\end{tabular}                                                       & RankScr: 0.724 \\ \midrule
    \textbf{Ours GLM (2)}      & \begin{tabular}[c]{@{}l@{}}Remove book3 from book1, Take book2 from table,\\ Place book2 on book3, Take book1 from table,\\ Place book1 on book2\end{tabular}                                                       & \textbf{RankScr: 0.788} \\ \bottomrule
    \end{tabular}%
    }
\end{table}

\subsection{Failure Case Analysis}
\textbf{Schema Set with No Plan Found:}\quad We encountered instances where no solvable action schema set was generated, primarily due to limitations in the LLM's reasoning capabilities. The use of open-source LLMs, while more accessible, may result in a lower success rate compared to more advanced proprietary models like GPT-4o. Specifically, with 7 LLM instances, we observed occasional failures of generating solvable sets action schemas for the \emph{libraryworld} and \emph{minecraft} domains. Nevertheless, solvable schema sets were consistently obtained across all domains when the number of LLM instances was increased to 10 (see Appendix~\ref{app:extra_results} for a breakdown of schema set yield by LLM instance count).\todo[inline, color=green]{tell readers to refer to the appendix for more details}

\noindent \textbf{Unexpected Preference:}\quad In the \emph{Dungeon} domain, human assessors unexpectedly preferred ToT-generated plans over both the reference plan and the proposed pipeline's plans. Further analysis revealed that the ToT plans consistently included a step: \emph{grabbing a sword}. Interestingly, grabbing a sword was not a necessary step for solving the given problem. Consequently, symbolic planners, focused on optimal pathfinding, excluded this step from their plans. However, this ``unnecessary'' step of acquiring a sword aligns with common strategies in Dungeon games, where players typically prioritize preparedness. Thus, this action strongly appealed to human assessors, causing them to rank the ToT-generated plans higher.

\section{Conclusion}
Existing methods for generating action schemas and plans in LLM-symbolic planning systems typically provide users with a single option. This limitation stems from the heavy reliance on expert intervention, which creates an efficiency bottleneck and risks introducing biased interpretations that misinterpret the true intention of a task. To address these challenges, we propose a novel 3-step pipeline that generates multiple action schema and plan candidates, offering users a range of ranked options to choose from. Our findings demonstrate that a full end to end LLM-symbolic planner is possible without expert intervention, paving the way for democratizing planning systems for a broader audience. One limitation in this work is the lack of direct evaluation methods for assessing the quality of generated action schema sets. Metrics like ``bisimulation'' \citep{coulter2022theory} or ``heuristic domain equivalence'' \citep{oswald2024large} require the generated schema sets to have the same action parameters as a predefined reference set. This approach doesn't suit our context, where action parameters are flexible and inferred in real-time from natural language descriptions. This highlights the need for new evaluation metrics suited to such dynamically generated action schema models.

\bibliography{aaai25}

\onecolumn
\pagebreak
\twocolumn
\appendix

\section{Probabilistic Analysis for Action Schema Combinations}
\label{app:pool_analysis}
Let's consider a domain consisting of $M$ actions, where the probability of generating a correct action schema for each action is $p$. The probability of obtaining a solvable (i.e., correctly formed and usable by the symbolic planner) set of action schemas for the domain is then $p^M$. $p^M$ diminishes exponentially as $M$ increases. For instance, with $p=0.05$ (based on observations from \citet{guan2023leveraging}) and $M=5$, the probability of obtaining a solvable set of action schemas from vanilla LLMs is only 0.00003125\%. In contrast, by combining the action schemas generated by each LLM instance, we can obtain approximately $\binom{N}{1}^M$ different sets of action schemas, resulting in a vast number of action schema set candidates.

\noindent \textbf{Success Analysis:}\quad Assume each action schema $\alpha_{ij}$ has an independent probability $p_{ij}$ of being solvable. For simplicity, let's assume $p_{ij} = p$ for all $i$ and $j$. A set of action schemas in the PDDL domain model is considered as solvable if all its action schemas are solvable. Therefore, the probability that a generated set of action schemas is not solvable is $(1-p^M)$, and the probability that none of the generated sets are solvable is $(1-p^M)^{N^M}$ and the probability of at least one combination is solvable is $1 - (1-p^M)^{N^M}$. For large $N$, as long as $p \in (0,1)$, $(1-p^M)^{N^M}$ approaches $0$. For example, with 10 LLM instances, a domain of 5 actions, and a solvability probability of $0.05$, the probability of obtaining at least one solvable set of action schemas is $1 - (1 - (0.05)^{5})^{5^{10}} = 1 -  0.0484 = 95.2\%$.

\section{Conformal Prediction Details}
\label{app_sec:cp_algo}

With a user-specified confidence level $1-\epsilon$, we calculate the $\frac{\lceil (n + 1)(1 - \epsilon) \rceil}{n}$ empirical quantile of the cosine similarity scores for true positive pairs in the calibration set, denoted as $\hat q$, where $n$ is the size of the calibration data pairs. Statistically, it ensures that at least $1-\epsilon$ fraction of the true positive pairs will have a cosine similarity score greater than $\hat q$. We then use $\hat q$ as the threshold to filter out action schemas with cosine similarity scores below this value. Statistically, this approach guarantees a high probability of preserving true positive action schema candidates while minimizing the size of the filtered set \citep{sadinle2019least}. For an illustration of this process, see \emph{step 2} in Figure~\ref{fig:overview_our_pipeline}. The pseudo-code for calculating the empirical quantile is shown in Algorithm~\ref{app:cpq_algo}.

During our experiments, we set $\epsilon = 0.2$ to ensure a high confidence level while maintaining a reasonable number of solvable action schema candidates. This choice was based on the trade-off between the confidence level and the number of solvable action schema candidates generated. A higher $\epsilon$ leads to a more stringent filtering process, resulting in a smaller number of solvable action schema candidates so as to save computational resources. In our experiments, the pretrained model \textit{all-roberta-large-v1}'s $\hat q$ value at $\epsilon = 0.2$ was 0.398. Thus, any action schema with a cosine similarity score below 0.398 was filtered out, ensuring that only high-quality action schemas were passed to the symbolic planner.

\begin{algorithm}[H]
    \caption{Calculate Empirical Quantile ($\hat{q}$)}
    \label{app:cpq_algo}
    \begin{algorithmic}[1] 
        \Require Calibration set $\{\mathcal Z(\alpha), \alpha\}_n$, where $\mathcal Z(\alpha)$ is natural language description of the corresponding true action schema $\alpha$;

        $n$ is the size of the calibration set;

        Significance level $\epsilon$;

        Sentence encoder model $E$

        \LComment{Obtain the Cosine Similarity Score}

        \State $\{e\}_n \gets \left\{\frac{E(\mathcal Z( \alpha)) \cdot E( \alpha)}{\|E(\mathcal Z( \alpha))\| \|E( \alpha)\|}\right\}_n$
        \LComment{Compute quantile level}
        \State $q_\text{level} \gets \frac{\lceil (n-1) \times (1-\epsilon) \rceil}{n}$
        \LComment{Compute empirical quantile}
        \State $\hat q \gets \textit{np.quantile($\{e\}_n$, $q_\text{level}$, method=`lower')}$ 
        \State \Return $\hat q$
    \end{algorithmic}
\end{algorithm}

\section{Domain Leakage Investigation Methodology}
\label{app:leak_investigation_methodology}
To determine whether the testing PDDL domains had been leaked to the training data of the large language models (LLMs), we conducted a specific investigation. This involved providing the LLMs with partial information about the PDDL domains, specifically the types, predicates, and the first action schema. Following this, we asked the LLMs to generate the remaining content of the domain.

In our findings, the LLMs were able to accurately generate the complete action schemas for the well-known \emph{blocksworld}, \emph{tyreworld}, and \emph{logistics} environments, which have been extensively used in prior research, indicating that LLMs had likely been exposed to these domains during pretraining. In contrast, for the testing domains used in our experiments, namely \emph{minecraft}, \emph{dungeon}, and \emph{libraryworld}, the LLMs were unable to reconstruct the complete action schemas based on the partial information provided, thereby ensuring the integrity of our experimental conditions. 

\section{Symbolic Planner Details}
\label{app:symbolic_planner_details}
Modern symbolic planners can efficiently verify the solvability of a generated schema set by leveraging its ability to prove delete-free reachability in polynomial time \citep{bonet2001planning}. This leads to very quick verification of whether the generated schema set is viable for generating a plan. Additionally, modern planners can run efficiently on a single CPU thread, and multiple problem instances can be solved in parallel if more CPUs are available. For instance, an AMD Ryzen 5900 with 32 threads can easily check the solvability of up to 20,000 generated schema sets within 2 minutes. By applying a Conformal Prediction (CP) filtering mechanism, we can further reduce the number of schema sets that need to be checked, allowing us to verify solvability within seconds, as most unsuitable sets are filtered out.
\onecolumn

\section{Experimental Setup Details}
\label{app:experimental_setup}
This section consists of the following:
\begin{itemize}
    \item \S \ref{app:testing_tasks_specifications}: Detailed natural language descriptions and reference PDDL models for the testing domains \emph{Libraryworld} and \emph{Dungeon}.
    \item \S \ref{app:costar_framework}: Details on the CO-STAR prompt engineering framework used for LLM prompt engineering.
    \item \S \ref{app:prompt_template_for_action_schema}: Prompt template used for generating action schemas, including how to structure few-shot learning examples in the prompt and how to obtain CoT reasoning examples automatically from advanced LLMs. 
    \item \S \ref{app:prompt_template_for_tot_direct_planning}: Prompt template used for Tree-of-Thought direct LLM-based planning.
    \item \S \ref{app:syntax_correction}: Syntax correction process for the generated action schemas.
    \item \S \ref{app:model_training_configs}: LLM model configurations and training configurations for the sentence encoder model.
\end{itemize}

\lstdefinestyle{mystyle}{
    backgroundcolor=\color{backcolour},   
    commentstyle=\color{codegreen},
    keywordstyle=\color{magenta},
    numberstyle=\tiny\color{codegray},
    stringstyle=\color{codepurple},
    basicstyle=\ttfamily\footnotesize,
    breakatwhitespace=false,         
    breaklines=true,                 
    captionpos=b,                    
    keepspaces=true,                 
    numbers=none,                    
    showspaces=false,                
    showstringspaces=false,
    showtabs=false,                  
    tabsize=1,
    xleftmargin=0pt, 
}

\lstset{style=mystyle}

\subsection{Testing Tasks Specifications}
\label{app:testing_tasks_specifications}

\begin{lstlisting}[language=domain_description, caption=Libraryworld Domain Descriptions, label=lst:libworld_domain_desc]
Domain Description: This domain is structured to allow organizing and managing books within a library setting. The actions and predicates support the movement of books between tables and shelves, ensuring that conditions like accessibility and the librarian's hands being free are met. Additionally, it includes managing book categories, shelf space, and check-out/return processes to reflect a more complex library system.

Action Description:
take-from-table:
    detailed: Imagine you're a librarian managing a table full of books. The 'take-from-table' action allows you to pick up a book that is on the table, provided it is accessible and your hands are free. This action simulates the scenario where you find a book on the table, ensure it's not covered by any other book, and then pick it up, thus holding it in your hands.
    ambiguous: Pick up a book from the table if it's not covered and your hands are empty.
place-on-shelf:
    detailed: Consider a librarian holding a book and standing near a shelf. The 'place-on-shelf' action involves placing the held book on top of another book on the shelf, given that the book on the shelf is accessible. This action results in the held book becoming accessible, the book on the shelf becoming inaccessible, and the librarian's hands becoming free.
    ambiguous: Put a book you're holding on top of another accessible book on the shelf.
...
Predicate List:
(on-shelf ?x ?y - book) ;; ?x is on top of ?y on the shelf
(on-table ?x - book) ;; ?x is on the table
(accessible ?x - book) ;; ?x is accessible (not covered)
(hands-free) ;; The hands of the librarian are free
(holding ?x - book) ;; The librarian is holding ?x
(belongs-to-category ?x - book ?cat - category) ;; ?x belongs to the category ?cat
(shelf-empty ?cat - category) ;; The shelf for category ?cat is empty
(shelf-overflow ?cat - category) ;; The shelf for category ?cat is full
(book-request ?book - book) ;; There is a request for book ?book
(return-due ?book - book) ;; Book ?book is due for return
(checked-out ?book - book) ;; Book ?book is checked out
\end{lstlisting}

\begin{lstlisting}[language=problem_description, caption=Libraryworld Problem Descriptions, label=lst:libworld_problem_desc]
Initial State Description: In the library, there are three books: Book1, Book2, and Book3. Book3 is on top of Book1 and they are both on the shelf, while Book2 is on the table. Book1 can also be considered as on the table it is just at the bottom of the shelf. Both Book2 and Book3 are accessible, meaning they can be interacted with. The library worker's hands are free. Book1 belongs to the Fiction category, Book2 belongs to the NonFiction category, and Book3 belongs to the Reference category.

Goal State Description: The goal is to have Book2 on top of Book3, and also Book1 on top of Book2.
\end{lstlisting}

\begin{figure}[H]
    \centering
    \begin{minipage}[t]{0.45\textwidth}
        \centering
        \begin{lstlisting}[language=pddl, caption={Libraryworld Reference PDDL Domain Model}, label=lst:libworld_pddl_domain]
(define (domain libraryworld)
(:requirements :strips :typing :negative-preconditions)
(:types book category)
(:predicates
    (on-shelf ?x ?y - book)
    (on-table ?x - book)
    (accessible ?x - book)
    (hands-free)
    ...
)
(:action take-from-table
    :parameters (?x - book)
    :precondition (and (accessible ?x) (on-table ?x) (hands-free))
    :effect (and (not (on-table ?x))
                (not (accessible ?x))
                (not (hands-free))
                (holding ?x))
)
(:action place-on-table
    :parameters (?x - book)
    :precondition (holding ?x)
    :effect (and (not (holding ?x))
                (accessible ?x)
                (hands-free)
                (on-table ?x))
)
...
)
            \end{lstlisting}
    \end{minipage}
    \hspace{0.5cm}
    \begin{minipage}[t]{0.45\textwidth}
        \centering
        \begin{lstlisting}[language=pddl, caption={Libraryworld Reference PDDL Problem Model}, label=lst:libworld_pddl_problem]
(define (problem organize-books)
(:domain libraryworld)
(:objects
    Book1 Book2 Book3 - book
    Fiction Non_Fiction Reference - category
)
(:init
    (on-table Book1)
    (on-shelf Book3 Book1)
    (on-table Book2)
    (accessible Book2)
    (accessible Book3)
    (hands-free)
    (belongs-to-category Book1 Fiction)
    (belongs-to-category Book2 Non_Fiction)
    (belongs-to-category Book3 Reference)
    ...
)
(:goal
    (and
        (on-shelf Book2 Book3)
        (on-shelf Book1 Book2)
    )
)
)
        \end{lstlisting}
    \end{minipage}
\end{figure}

\begin{lstlisting}[language=domain_description, caption=Dungeon Domain Descriptions, label=lst:dungeon_domain_desc]
Domain Description: Help the hero to get out of dungeon! A hero woke up in a dungeon full of monsters and traps (perhaps the party last night went wrong...) and needs your help to get out.  Here are basic facts for the dungeon domain: - The dungeon contains rooms that are **connected** by corridors (dungeon can thus be represented by undirected graph) - each room can be **empty**, or can have a **monster** in it, or can have a **trap** in it, or can have a **sword** in it - one of the empty rooms is the **goal**: it has an exit, so the hero can escape.

Action Description:
move:
    detailed: The hero can **move** to an adjacent room (connected by a corridor) that has not been destroyed (i.e., the hero has not already visited the room). When this action is executed, the original cell get destroyed.
    ambiguous: Hero can move if the - hero is at current location - cells are connected, - there is no trap in current loc, and - destination does not have a trap/monster.
pick-sword:
    detailed: **Pickup** the sword if present in the room the hero is currently in and the hero is empty handed.
    ambiguous: Hero picks a sword if he's in the same location.
...
Predicate List:
(at-hero ?loc - cells) ;;  Hero's cell location
(at-sword ?s - swords ?loc - cells) ;; Sword cell location
(has-monster ?loc - cells) ;; Indicates if a cell location has a monster
(has-trap ?loc - cells) ;; Indicates if a cell location has a trap
(is-destroyed ?obj) ;; Indicates if a chell or sword has been destroyed
(connected ?from ?to - cells) ;; connects cells
(arm-free) ;; Hero's hand is free
(holding ?s - swords) ;; Hero's holding a sword
(trap-disarmed ?loc) ;; It becomes true when a trap is disarmed
\end{lstlisting}
    
\begin{lstlisting}[language=problem_description, caption=Dungeon Problem Descriptions, label=lst:dungeon_problem_desc]
Initial State Description: In the dungeon, the hero starts at cell5, with free hands ready for action. The hero is aware of the dungeon's layout, which consists of multiple interconnected cells. A sword is located in cell4. The dungeon contains dangerous monsters located in cell3 and cell8, and a trap is present in cell2. The hero must navigate this treacherous environment, using the connections between the cells to move around. The connections are as follows:

    Cell1 is connected to cell2.
    Cell2 is connected to cell1 and cell3.
    Cell3 is connected to cell2 and cell4.
    Cell4 is connected to cell3 and cell5.
    Cell5 is connected to cell4 and cell8.
    Cell6 is connected to cell7.
    Cell7 is connected to cell6 and cell8.
    Cell8 is connected to cell7 and cell5.
    Cell2 is also connected to cell6.
    Cell3 is also connected to cell7.
    Cell4 is also connected to cell8.

Goal State Description: The hero's ultimate objective is to reach cell1 safely.
\end{lstlisting}

\begin{figure}[H]
    \centering
    \begin{minipage}[t]{0.45\textwidth}
        \centering
        \begin{lstlisting}[language=pddl, caption={Dungeon Reference PDDL Domain Model}, label=lst:dungeon_pddl_domain]
(define (domain rpggame)
(:requirements :typing :negative-preconditions
)
(:types
    swords cells
)
(:predicates
    (at-hero ?loc - cells)
    (at-sword ?s - swords ?loc - cells)
    (has-monster ?loc - cells)
    (has-trap ?loc - cells)
    ...
)
(:action move
    :parameters (?from ?to - cells)
    :precondition (and
        (connected ?from ?to)
        (at-hero ?from)

        (not (has-trap ?from))
        (not (is-destroyed ?to))
        (not (has-trap ?to))
        (not (has-monster ?to))

    )
    :effect (and
        (at-hero ?to)
        (is-destroyed ?from)
        (not (at-hero ?from))
    )
)

...
)
            \end{lstlisting}
    \end{minipage}
    \hspace{0.5cm}
    \begin{minipage}[t]{0.45\textwidth}
        \centering
        \begin{lstlisting}[language=pddl, caption={Dungeon Reference PDDL Problem Model}, label=lst:dungeon_pddl_problem]
(define (problem p1-dangeon)
(:domain rpggame)
(:objects
    cell1 cell2 cell3 cell4 cell5 cell6 cell7 cell8 - cells
    sword1 - swords
)
(:init

    ;Initial Hero Location
    (at-hero cell5)
    ;He starts with a free arm
    (arm-free)
    ;Initial location of the swords
    (at-sword sword1 cell4)
    ;Initial location of Monsters
    (has-monster cell3)
    (has-monster cell8)
    ;Initial location of Traps
    (has-trap cell2)
    
    ;Graph Connectivity
    (connected cell1 cell2)
    (connected cell2 cell1)
    (connected cell2 cell3)
    (connected cell3 cell2)
    (connected cell3 cell4)
    (connected cell4 cell3)
    (connected cell4 cell5)
    ...
)
(:goal (and
            (at-hero cell1)
))

)
        \end{lstlisting}
    \end{minipage}
\end{figure}

\twocolumn

\subsection{CO-STAR Framework}
\label{app:costar_framework}
The CO-STAR framework is a structured template for crafting effective prompts for LLMs. Developed by GovTech Singapore's Data Science and Artificial Intelligence Division, CO-STAR is designed to improve the quality of LLM-generated responses by systematically addressing key aspects that influence output.

The CO-STAR acronym stands for:
\begin{itemize}
    \item \textbf{Context (C)}: Provide background information on the task.
    \item \textbf{Objective (O)}: Define the specific task you want the LLM to perform.
    \item \textbf{Style (S)}: Specify the desired writing style for the LLM's response.
    \item \textbf{Tone (T)}: Set the attitude or emotional quality of the response.
    \item \textbf{Audience (A)}: Identify the intended recipients of the response.
    \item \textbf{Response (R)}: Outline the expected format of the response.
\end{itemize}

\subsection{Prompt Template for Action Schema Generation and Obtaining CoT Reasoning Examples}
\label{app:prompt_template_for_action_schema}

Based on the CO-STAR framework, we designed a structured prompt template for generating action schemas. The template includes the following components:
\vspace{0.5cm}
\begin{lstlisting}[language=chat]
System: # CONTEXT #
You are a tool called PDDL Modeling Assistant. \
You are a technical experts in constructing Planning Domain Definition Language (PDDL) models via the natural language context.

# OBJECTIVE #
* Construct parameters, preconditions and effects based on the domain information, action description and the action name.
* All variables in the preconditions and effects must be listed in the action's parameters. This restriction helps maintain the action's scope and prevents ambiguity in the planning process.
* Do not use predicates that are not defined in the available predicates list to construct the preconditions and effects.
* When the natural language description is ambiguous or certain predicate changes are implied, make reasonable assumptions based on common sense to fill up the implicit predicate in the PDDL action.

# STYLE #
Follow the writing style of technical experts. The output can be parsed by a machine, so it is important to follow the structured format.

# TONE #
Be precise and concise in constructing the PDDL action. The PDDL action should be clear and unambiguous.

# AUDIENCE #
Your audience is someone who is trying to learn how to construct PDDL actions from natural language descriptions.

# RESPONSE #
The response should be in the following format:
---
**Explanation:** [Your explanation here]

**Response:**
Parameters:
1. ?x - [type]: [parameter description]
2. ...

Preconditions:
```
(and
    ([predicate_1] ?x)
)
```

Effects:
```
(and
    (not ([predicate_2] ?x))
    ([predicate_2] ?x)
    ...
)
```
---
\end{lstlisting}

We further include two example query-answer pairs to facilitate in-context learning for LLMs. The example are taken from the training domains rather than the testing domains to ensure the integrity of the experimental conditions.

\begin{lstlisting}[language=chat]
Example Human: One or two examples from other domains for illustrating the input and output formats.
Here are two examples from the newspapers domain for demonstrating the output format.
        
Domain information: This domain is structured to model a simple newspaper delivery system, where the main tasks involve picking up newspapers from a home base, moving to different locations, and delivering the papers to locations that require them.

A list of available predicates
1. (at ?loc - loc) ;; the agent is at loc location
2. (is_Home_Base ?loc - loc) ;; the location is the home base
3. (satisfied ?loc - loc) ;; the location has received the newspaper
4. (wants_Paper ?loc - loc) ;; the location needs a newspaper
5. (unpacked ?paper - paper) ;; the paper is unpacked
6. (carrying ?paper - paper) ;; the agent is carrying the paper
Example 1
Action Description: RoboDelivery is at the home base (loc_home).
There is an unpacked newspaper (paper1) at the home base.
RoboDelivery performs the pick-up action.

    Preconditions: RoboDelivery is at loc_home (which is the home base), and paper1 is unpacked.
    Effects: RoboDelivery is now carrying paper1, and paper1 is no longer unpacked.

Action name: pick-up


Your answer:
---

Example AI: **Explanation:**

Let's imagine a situation where someone, like a delivery person, needs to pick up newspapers from a place, like a home base, and deliver them to various locations. This situation can be broken down into smaller, specific steps to make it easy for a computer to understand and execute. This is where the action schema you mentioned comes into play.

The action is called "pick-up," and it involves the delivery person picking up newspapers to deliver. This action is defined in a language called PDDL, which is like giving a computer a recipe to follow.

1. **Parameters**: These are like ingredients in a recipe. Here, the ingredients are any newspaper (`?paper`) and a location (`?loc`), which in this scenario is the home base.

2. **Preconditions**: These are conditions that must be true before the action can start. Think of it like checking if you have all your ingredients before you start cooking.
   - `(at ?loc)`: The delivery person must be at the location specified (`?loc`). This makes sure they are actually at the home base to pick up the newspapers.
   - `(is_Home_Base ?loc)`: The location where they are must be the home base. This is crucial because newspapers ready to be delivered are stored here.
   - `(unpacked ?paper)`: The newspapers need to be unpacked and ready to be picked up. It ensures the newspapers are not in boxes or still bundled up, making them easy to grab and go.

3. **Effects**: These describe what happens after the action is completed.
   - `(not (unpacked ?paper))`: The newspaper is no longer in an unpacked state at the base because it has been picked up.
   - `(carrying ?paper)`: The newspaper is now being carried by the delivery person, ready to be delivered to someone who needs it.

This schema helps the computer understand exactly what needs to happen when the "pick-up" action is called in the program. By specifying preconditions, we ensure the action makes sense (you can't pick up a newspaper if you're not there or if the papers aren't ready). The effects tell the computer how the world changes after the action, like the newspaper moving from being unpacked at the base to being carried for delivery.

By breaking down tasks this way, we help the computer efficiently and correctly handle tasks that seem simple to us but are quite detailed when programmed.
                
**Response:**
Parameters:
1. ?from - loc: [short description of the parameter]
2. ?to - loc: [short description of the parameter]

Preconditions:
```
(at ?from)
```

Effects:
```
(and
    (not (at ?from))
    (at ?to)
)
```

---
\end{lstlisting}

As demonstrated in the example above, we include a detailed explanation section. This approach, known as Chain-of-Thought (CoT), has been shown to enhance LLMs' reasoning capabilities by explicitly prompting them to show intermediate steps in their reasoning process. 

The following section presents the actual query-response pairs generated by the LLMs during the action schema generation process.
\vspace{0.5cm}

\begin{lstlisting}[language=chat]
System: [SAME AS THE PREVIOUS EXAMPLE]
Human: Question: Here is the task.
A natural language description of the domain
Domain information: This domain is structured to allow organizing and managing books within a library setting. The actions and predicates support the movement of books between tables and shelves, ensuring that conditions like accessibility and the librarian's hands being free are met. Additionally, it includes managing book categories, shelf space, and check-out/return processes to reflect a more complex library system.

A list of available predicates
1. (on-shelf ?x ?y - book) ;; ?x is on top of ?y on the shelf
2. (on-table ?x - book) ;; ?x is on the table
3. (accessible ?x - book) ;; ?x is accessible (not covered)
4. (hands-free) ;; The hands of the librarian are free
5. (holding ?x - book) ;; The librarian is holding ?x
6. (belongs-to-category ?x - book ?cat - category) ;; ?x belongs to the category ?cat
7. (shelf-empty ?cat - category) ;; The shelf for category ?cat is empty
8. (shelf-overflow ?cat - category) ;; The shelf for category ?cat is full
9. (book-request ?book - book) ;; There is a request for book ?book
10. (return-due ?book - book) ;; Book ?book is due for return
11. (checked-out ?book - book) ;; Book ?book is checked out

Action Description: Imagine you're a librarian managing a table full of books. The 'take-from-table' action allows you to pick up a book that is on the table, provided it is accessible and your hands are free. This action simulates the scenario where you find a book on the table, ensure it's not covered by any other book, and then pick it up, thus holding it in your hands.

Action name: take-from-table


Your answer:

AI: **Explanation:**
In the context of a library, the "take-from-table" action represents the physical act of a librarian picking up a book from a table. This action is only feasible under certain conditions, which are represented by the preconditions of the action. The effects describe the outcome or the change in the state of the world once the action is executed.

The parameters of the action are the book that is to be picked up and potentially the table it is on, though the table is not explicitly mentioned in the available predicates, so we'll focus on the book.

The preconditions ensure that:
1. The book is on the table and accessible, meaning it's not obscured or covered by other books.
2. The librarian's hands are free, indicating that they are not currently holding any other book.

Once these conditions are met, the librarian can perform the "take-from-table" action, which results in two main effects:
1. The book is no longer on the table, as it is now being held by the librarian.
2. The librarian is no longer hands-free, as they are now holding the book.

This action is a fundamental part of managing books within the library, allowing the librarian to move books from tables to shelves or to handle checkouts and returns.

**Response:**
Parameters:
1. ?book - book: [The book that the librarian will pick up from the table]

Preconditions:
```
(and
    (on-table ?book)
    (accessible ?book)
    (hands-free)
)
```

Effects:
```
(and
    (not (on-table ?book))
    (not (hands-free))
    (holding ?book)
)
```
\end{lstlisting}

We can see that the LLM-generated response closely aligns with the expected format, demonstrating the effectiveness of the prompt template in guiding the LLM to generate accurate action schemas.

\subsection{Prompt Template for ToT Direct Planning}
\label{app:prompt_template_for_tot_direct_planning}

The Tree-of-Thought (ToT) direct planning approach uses a prompt template similar in structure to that used for action schema generation. The key distinction lies in the system prompt, which instructs the LLM to generate plans rather than action schemas. The system prompt for ToT direct planning is as follows:
\begin{lstlisting}[language=chat]
System: # CONTEXT #
You are a tool called Automated Planning Action Planner. 
You are a technical expert in constructing and evaluating the quality of action plans via the natural language context.
# OBJECTIVE #
* Add only one more action to the given action plan sequence towards achieving the goal state.
* Provide the action name and the objects that this action will interact with.
* Evaluate the confidence score that continuing with the suggested action plan will eventually lead to the goal state.
* Determine if the action plan has already reached the goal state.
* Output only one action at each step and stop.
* You should think step by step, you think think more steps. 
...
# RESPONSE #
The response should be in the following format:
---
**Response:**
1. Action {n}: [Action Name]
2. Objects: 
   - ?x - [type]: [object description]
   - ...
   
3. Updated Action Plan: 
   - [Action 1 and the objects involved] 
   - [Action 2 and the objects involved]
   - ...
   - [Action {n} and the objects involved]

**Confidence Evaluation:**
[Analyze the updated action plan, then at the last line conclude "The confidence score is {s}", where s is an integer from 1 to 10]

**Goal State Check:**
[Briefly analyze the current state, then at the last line conclude "The planning is continuing" or "The planning is completed"]
---
\end{lstlisting}

\subsection{Syntax Correction Process}
\label{app:syntax_correction}
After obtaining the raw outputs from the LLMs, our pipeline post-processes these outputs into structured PDDL snippets, correcting any potential syntax errors present in the initial generation. Although previous research has demonstrated that LLMs can effectively correct syntax errors when given feedback from validation tools \citep{guan2023leveraging, roziere2023code, Silver2023GeneralizedPI}, this approach is computationally expensive because it requires numerous iterative calls to the LLM.

As our work primarily focuses on addressing semantic (factual) errors in LLM outputs, we opted for a more efficient approach to syntax correction. Instead of using LLMs, we employed the PDDL parser tool from \url{https://github.com/jan-dolejsi/vscode-pddl} to directly correct syntax errors. This tool specifically addresses syntax errors, thus it will not introduce any semantic changes to the generated PDDL snippets.

\subsection{Model and Training Configurations}
\label{app:model_training_configs}

LLM configuration is as follows:

\begin{table}[H]
    \caption{LLM model configuration}
    \label{tab:llm-model-config}
    \resizebox{\columnwidth}{!}{%
    \begin{tabular}{@{}lcc@{}}
    \toprule
    Parameter          & Action Schema Generation & Direct Plan Generation                                                                     \\ \midrule
    model\_name        & glm-4-0520               & \begin{tabular}[c]{@{}c@{}}glm-4-0520,\\ gpt-4o-2024-05-13,\\ gpt-3-turbo\end{tabular}     \\ \midrule
    top\_p             & 0.3                      & 0.80                                                                                       \\ \midrule
    temperature        & 0.3                      & 0.99                                                                                       \\ \midrule
    max\_tokens        & 1024                     & 1024                                                                                       \\ \midrule
    tree\_breadth      & -                        & 3                                                                                          \\ \midrule
    tree\_depth\_ratio & -                        & \begin{tabular}[c]{@{}c@{}}1.5\\ tree depth = ratio *\\ referece plan length)\end{tabular} \\ \bottomrule
    \end{tabular}%
    }
\end{table}

The finetuning configuration for the sentence encoder model is as follows:

\begin{table}[H]
    \caption{Sentence Encoder Finetuning Configs}
    \centering
    \label{tab:sentence_enc_finet_config}
    \resizebox{0.75\columnwidth}{!}{%
    \begin{tabular}{@{}ll@{}}
    \toprule
    Parameter                & Value                                                                                                                                   \\ \midrule
    train\_negative\_weights & \begin{tabular}[c]{@{}l@{}}{[}0.0, 0.4, 0.6{]}\\ (ratio of selecting easy,\\ semi-hard, hard negatives\\ from the training dataset)\end{tabular} \\ \midrule
    train\_batch\_size       & 256                                                                                                                                     \\ \midrule
    training\_epoch          & 40                                                                                                                                      \\ \midrule
    sentence\_encoder        & all-roberta-large-v1                                                                                                                    \\ \bottomrule
    \end{tabular}%
    }
\end{table}
\noindent \textbf{Finetuning Details:}\quad The sentence encoder model was fine-tuned on a dataset of 200,000 synthesized samples generated by action schema manipulation. The process utilized an NVIDIA A100 80GB PCIe GPU on a Linux 5.14.0 system. It takes about 11 hours to complete the training process. More details can be found in the code repository.

\onecolumn
\section{Extra Results}
\label{app:extra_results}
\begin{figure}[H]
    \centering
    \includegraphics[width=0.6\columnwidth]{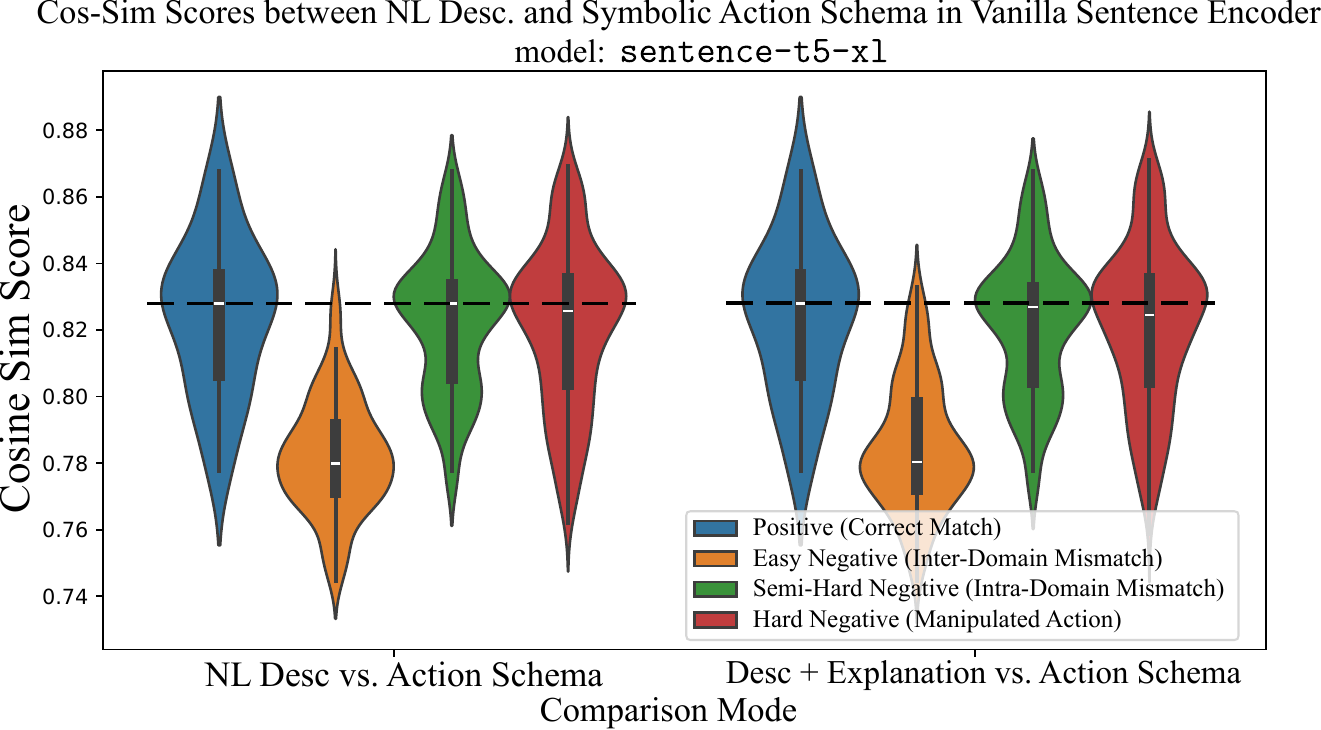}
    \caption{The pre-trained \texttt{sentence-t5-xl} sentence encoder demonstrates semantic alignment for matched action schemas and misalignment for mismatched ones, supporting the concept of semantic equivalence across different representations of the same entity.}
    \label{fig:cosine_sim_comparison_before_finetune_t5_xl}
\end{figure}

\begin{table}[H]
    \caption{Detailed action schema generation results of the proposed pipeline without CP filtering}
    \label{tab:detailed_out_pipe_result_without_cp}
    \resizebox{\columnwidth}{!}{%
    \begin{tabular}{lcccccccc}
        \hline
        Domain Name  & Desc Granularity & LLM instance \# & Total Combinations & Solved Combinations & Distinct Plan \# & Avg. Plan Length & Applied CP Threshold & CP Threshold Value \\ \hline
        libraryworld & detailed         & 7               & 18144              & 0                   & N/A              & N/A              & False                & N/A                \\
        libraryworld & detailed         & 10              & 86400              & 4560                & 7                & 6.57             & False                & N/A                \\
        libraryworld & ambiguous        & 7               & 26460              & 0                   & N/A              & N/A              & False                & N/A                \\
        libraryworld & ambiguous        & 10              & 124416             & 17976               & 7                & 5.14             & False                & N/A                \\ \hline
        dungeon      & detailed         & 7               & 600                & 600                 & 5                & 4                & False                & N/A                \\
        dungeon      & detailed         & 10              & 2800               & 2099                & 5                & 4                & False                & N/A                \\
        dungeon      & ambiguous        & 7               & 360                & 180                 & 1                & 5                & False                & N/A                \\
        dungeon      & ambiguous        & 10              & 700                & 140                 & 1                & 5                & False                & N/A                \\ \hline
        minecraft    & detailed         & 7               & 1323               & 216                 & 2                & 3                & False                & N/A                \\
        minecraft    & detailed         & 10              & 5250               & 3598                & 4                & 3                & False                & N/A                \\
        minecraft    & ambiguous        & 7               & 7350               & 2800                & 5                & 3                & False                & N/A                \\
        minecraft    & ambiguous        & 10              & 21600              & 6000                & 6                & 3                & False                & N/A                \\ \hline
        \end{tabular}%
        }
\end{table}

\begin{table}[H]
    \caption{Detailed action schema generation results of the proposed pipeline WITH CP filtering}
    \label{tab:pipeline_result_with_cp}
    \resizebox{\columnwidth}{!}{%
    \begin{tabular}{lcccccccc}
        \hline
        Domain Name  & Desc Granularity & LLM instance \# & Total Combinations & Solved Combinations & Distinct Plan \# & Avg. Plan Length & Applied CP Threshold & CP Threshold Value \\ \hline
        libraryworld & detailed         & 7               & 672                & 0                   & N/A              & N/A              & True                 & 0.398              \\
        libraryworld & detailed         & 10              & 2160               & 60                  & 2                & 5                & True                 & 0.398              \\
        libraryworld & detailed         & 15              & 2340               & 104                 & 2                & 5                & True                 & 0.398              \\
        libraryworld & ambiguous        & 7               & 252                & 0                   & N/A              & N/A              & True                 & 0.398              \\
        libraryworld & ambiguous        & 10              & 3240               & 560                 & 6                & 5.3              & True                 & 0.398              \\
        libraryworld & ambiguous        & 15              & 37440              & 852                 & 10               & 6.3              & True                 & 0.398              \\ \hline
        dungeon      & detailed         & 7               & 24                 & 24                  & 3                & 4                & True                 & 0.398              \\
        dungeon      & detailed         & 10              & 48                 & 36                  & 3                & 4                & True                 & 0.398              \\
        dungeon      & detailed         & 15              & 128                & 96                  & 1                & 4                & True                 & 0.398              \\
        dungeon      & ambiguous        & 7               & 18                 & 9                   & 1                & 5                & True                 & 0.398              \\
        dungeon      & ambiguous        & 10              & 20                 & 4                   & 1                & 5                & True                 & 0.398              \\
        dungeon      & ambiguous        & 15              & 168                & 108                 & 3                & 4                & True                 & 0.398              \\ \hline
        minecraft    & detailed         & 7               & 441                & 0                   & N/A              & N/A              & True                 & 0.398              \\
        minecraft    & detailed         & 10              & 945                & 630                 & 2                & 3                & True                 & 0.398              \\
        minecraft    & detailed         & 15              & 1287               & 720                 & 5                & 3                & True                 & 0.398              \\
        minecraft    & ambiguous        & 7               & 2940               & 1120                & 6                & 3                & True                 & 0.398              \\
        minecraft    & ambiguous        & 10              & 6720               & 1920                & 6                & 3                & True                 & 0.398              \\
        minecraft    & ambiguous        & 15              & 45760              & 16384               & 8                & 2.63             & True                 & 0.398              \\ \hline
        \end{tabular}%
        }
\end{table}

\textbf{Justification why tested with problems that only need short plans:}

Our focus on problems requiring short plans stems from two key reasons:
\begin{itemize}
    \item \textbf{Focus on Schema Set Construction:} The primary challenge in LLM-symbolic planning lies in accurately constructing the action schema set. Once this set is correctly defined, plan generation is handled by the symbolic planner, and the length of the plan becomes no more a crucial factor in evaluating the LLM-symbolic planning pipeline's performance. Our approach excels in generating accurate action schema sets, ensuring reliable plan generation regardless of length.
    \item \textbf{Limitations of Direct LLM Planning Models:} Direct LLM planning models like ``Tree of Thoughts'' (ToT) suffer from inherent limitations in long-term planning due to their probabilistic nature. Accuracy diminishes exponentially with each step. For example, even with a 99\% per-step accuracy, the probability of a correct 100-step plan plummets to 36.6\%. This makes direct LLM planners unsuitable for long-term planning. Thus, for fair comparison, we focus on the reasoning capabilities of different approaches under rational plan lengths.
\end{itemize} 

\noindent \textbf{Implication:}\quad Our justification highlights a significant advantage of the LLM-symbolic planning pipeline: the quality of plan generation is not affected by the length of the plan, but rather by the quality of the action schema set. This means that our pipeline can be generalized to plan generation of any length. In contrast, direct LLM planning models are fundamentally limited in their ability to guarantee soundness in long-term planning. 

\begin{table}[H]
    \caption{Detailed Plan for ToT direct planning models in Sussman Anomaly testing case}
    \label{tab:detailed_susman_ano_plan_tot}
    \resizebox{\columnwidth}{!}{%
    \begin{tabular}{@{}llc@{}}
    \toprule
    Model     & Plan                                                                                                                                                                                                                                                                                                           & Heuristic \\ \midrule
    ToT GLM   & {[}``take-from-table Book2", ``place-on-shelf Book2 Book3", ``take-from-table Book1", ``place-on-shelf Book1 Book2"{]}                                                                                                                                                                                             & 9.0       \\ \midrule
    ToT GLM   & {[}``take-from-table Book2", ``place-on-shelf Book2 onto Book3", ``take-from-table Book1", ``place-on-shelf Book1 onto Book2"{]}                                                                                                                                                                                   & 9.0       \\ \midrule
    ToT GLM   & {[}``take-from-table Book2", ``place-on-shelf Book2 onto Book3", ``remove-from-shelf Book1", ``place-on-shelf Book1 onto Book2"{]}                                                                                                                                                                                 & 9.0       \\ \midrule
    ToT GPT-3 & \begin{tabular}[c]{@{}l@{}}{[}``take-from-table Book2", ``place-on-shelf Book2 Book3", ``place-on-table Book1", "place-on-shelf Book1 Book2",\\ ``remove-from-shelf Book3", ``place-on-shelf Book3 Book2", ``place-on-shelf Book1 Book2", ``check-out Book1"{]}\end{tabular}                                          & 5.11      \\ \midrule
    ToT GPT-3 & \begin{tabular}[c]{@{}l@{}}{[}"take-from-table Book2", ``place-on-shelf Book2 Book3", ``place-on-table Book1", ``place-on-shelf Book1 Book2",\\ ``remove-from-shelf Book3", ``place-on-shelf Book3 Book2", ``remove-from-shelf Book2", ``place-on-shelf Book2 Book1"{]}\end{tabular}                                  & 5.11      \\ \midrule
    ToT GPT-3 & \begin{tabular}[c]{@{}l@{}}{[}``take-from-table Book2", ``place-on-shelf Book2 Book3", ``place-on-table Book1", ``place-on-shelf Book1 Book2",\\ ``place-on-shelf Book1 Book3", ``place-on-shelf Book2 Book1", ``place-on-shelf Book2 Book3", ``remove-from-shelf Book2", ``place-on-shelf Book2 Book1"{]}\end{tabular} & 4.11      \\ \midrule
    ToT GPT-4 & {[}``take-from-table (Book2)", ``place-on-shelf (Book2, Book3)", ``take-from-table (Book1)", ``place-on-shelf (Book1, Book2)"{]}                                                                                                                                                                                   & 8.5       \\ \midrule
    ToT GPT-4 & {[}``take-from-table(Book2)", ``place-on-shelf(Book2, Book3)", ``take-from-table(Book1)", ``place-on-shelf(Book1, Book2)"{]}                                                                                                                                                                                       & 8.5       \\ \midrule
    ToT GPT-4 & \begin{tabular}[c]{@{}l@{}}{[}``take-from-table(Book2)", ``remove-from-shelf(Book3)", ``place-on-table(Book3)", ``place-on-table(Book2)",\\ ``take-from-table(Book2)", ``place-on-table(Book2, Book3)", ``take-from-table(Book1)", ``place-on-table(Book1, Book2)", ``take-from-table(Book1)"{]}\end{tabular}           & 8.33      \\ \midrule
    ToT GPT-4 & \begin{tabular}[c]{@{}l@{}}{[}``take-from-table(Book2)", ``remove-from-shelf(Book3)", ``place-on-table(Book3)", ``place-on-table(Book2)",\\ ``take-from-table(Book2)", ``place-on-table(Book1)", ``take-from-table(Book3)", ``place-on-shelf(Book3, Book1)", ``take-from-table(Book1)"{]}\end{tabular}                  & 7.89      \\ \midrule
    ToT GPT-4 & \begin{tabular}[c]{@{}l@{}}{[}``take-from-table(Book2)", ``remove-from-shelf(Book3)", ``place-on-table(Book3)", ``place-on-table(Book2)",\\ ``take-from-table(Book3)", ``place-on-table(Book2)", ``place-on-table(Book3)", ``take-from-table(Book2)", ``place-on-shelf(Book2, Book3)"{]}\end{tabular}                   & 7.78      \\ \midrule
    ToT GPT-4 & \begin{tabular}[c]{@{}l@{}}{[}``take-from-table(Book2)", ``remove-from-shelf(Book3)", ``place-on-table(Book3)", ``place-on-table(Book2)",\\ ``take-from-table(Book3)", ``place-on-table(Book2)", ``place-on-table(Book3)", ``take-from-table(Book2)", ``place-on-table(Book3)"{]}\end{tabular}                          & 7.44      \\ \bottomrule
    \end{tabular}%
    }
\end{table}

\begin{table}[H]
    \caption{Detailed Plan for the proposed LLM-symbolic planning pipeline in Sussman Anomaly testing case}
    \label{tab:ourpipe_perf_in_sussman_detailed}
    \resizebox{\columnwidth}{!}{%
    \begin{tabular}{@{}llc@{}}
    \toprule
    Model    & Plan                                                                                                                                                                                                                                                                                                                                 & RankScr \\ \midrule
    Ours GLM & \begin{tabular}[c]{@{}l@{}}{[}``(remove-from-shelf book3 book1)", ``(take-from-table book2)", ``(place-on-shelf book2 book3)",\\ ``(take-from-table book1)", ``(place-on-shelf book1 book2)"{]}\end{tabular}                                                                                                                              & 0.788     \\ \midrule
    Ours GLM & \begin{tabular}[c]{@{}l@{}}{[}``(take-from-table book2)", ``(place-on-shelf book2 book3)", ``(remove-from-shelf book3 book1)",\\ ``(take-from-table book1)", ``(place-on-shelf book1 book2)"{]}\end{tabular}                                                                                                                              & 0.770     \\ \midrule
    Ours GLM & \begin{tabular}[c]{@{}l@{}}{[}``(remove-from-shelf book3 book1)", ``(take-from-table book1)", ``(place-on-shelf book1 book2)",\\ ``(take-from-table book2)", ``(place-on-shelf book2 book3)"{]}\end{tabular}                                                                                                                              & 0.724     \\ \midrule
    Ours GLM & \begin{tabular}[c]{@{}l@{}}{[}``(remove-from-shelf book3 book1 reference)", ``(take-from-table book2)", ``(place-on-shelf book2 book3)",\\ ``(take-from-table book1)", ``(place-on-shelf book1 book2)"{]}\end{tabular}                                                                                                                    & 0.632     \\ \midrule
    Ours GLM & \begin{tabular}[c]{@{}l@{}}{[}``(remove-from-shelf book3 book1)", ``(take-from-table book1)", ``(place-on-table book3)",\\ ``(check-out book1)", ``(take-from-table book2)", ``(place-on-table book1)",\\ ``(place-on-shelf book2 book3)", ``(take-from-table book1)", ``(return-book book1)", ``(place-on-shelf book1 book2)"{]}\end{tabular} & 0.569     \\ \midrule
    Ours GLM & \begin{tabular}[c]{@{}l@{}}{[}``(take-from-table book2)", ``(place-on-shelf book2 book3)", ``(remove-from-shelf book3 book1)",\\ ``(check-out book1)", ``(place-on-shelf book1 book2)"{]}\end{tabular}                                                                                                                                    & 0.552     \\ \midrule
    Ours GLM & \begin{tabular}[c]{@{}l@{}}{[}``(take-from-table book2 reference)", ``(place-on-shelf book2 book3)", ``(remove-from-shelf book3 book1 reference)",\\ ``(take-from-table book1 reference)", ``(place-on-shelf book1 book2)"{]}\end{tabular}                                                                                                & 0.512     \\ \bottomrule
    \end{tabular}%
    }
\end{table}

For more detailed generated action schema sets and plans, please go to the following folders in the code repository:
\begin{itemize}
    \item \texttt{data/07\_model\_output/tree\_of\_thought\_plans}
    \item \texttt{data/07\_model\_output/llm\_to\_domain\_to\_plans}
\end{itemize}

\end{document}